\documentclass{article}
\usepackage[utf8]{inputenc}
\usepackage{authblk}
\usepackage{main}
\usepackage{microtype}
\usepackage{graphicx}
\usepackage{subfig}
\usepackage{times}
\usepackage{latexsym}
\usepackage{amsmath,amsfonts,amssymb,amsthm}
\usepackage{float}
\usepackage{footnote}
\usepackage{enumitem}
\usepackage{bm}
\usepackage{arydshln}
\usepackage{booktabs}
\usepackage{multicol}
\usepackage{multirow}
\usepackage{color}
\usepackage{xcolor}     
\usepackage{colortbl}
\usepackage[table]{xcolor} 
\definecolor{royalblue}{RGB}{65,105,225}
\definecolor{purple}{RGB}{128,0,128}

\usepackage{bbding}
\usepackage{makecell}
\usepackage{mathtools}
\usepackage{imakeidx}
\usepackage{longtable}
\usepackage{wrapfig}
\makeindex
\usepackage{arydshln}
\usepackage{lipsum}
\usepackage{natbib}
\usepackage[edges]{forest}
\usepackage[normalem]{ulem}
\definecolor{mydarkblue}{rgb}{0,0.08,0.45}
\usepackage[colorlinks=true,linkcolor=mydarkblue,citecolor=mydarkblue,filecolor=mydarkblue,urlcolor=mydarkblue]{hyperref}
\usepackage{caption}
\usepackage{CJKutf8}
\usepackage{awesomebox} 
\usepackage{bbding}
\usepackage[most]{tcolorbox}

\usepackage[tikz]{bclogo}
\usepackage[framemethod=tikz]{mdframed}
\definecolor{bgblue}{RGB}{245,243,253}
\definecolor{ttblue}{RGB}{91,194,224}
\usetikzlibrary{positioning, shapes.geometric, shapes, arrows.meta}
\usepackage{tabularx}
\usepackage{geometry}
\usepackage{pifont}
\definecolor{myblue}{RGB}{0,80,160}
\definecolor{myyellow}{RGB}{210,170,0}
\definecolor{mygray}{RGB}{100,100,100}
\definecolor{mypurple}{RGB}{120,80,160}

\mdfdefinestyle{mystyle}{%
  rightline=true,
  innerleftmargin=10,
  innerrightmargin=10,
  outerlinewidth=3pt,
  topline=false,
  rightline=true,
  bottomline=false,
  skipabove=\topsep,
  skipbelow=\topsep
}

\newtcolorbox{myboxi}[1][]{
  breakable,
  title=#1,
  colback=red!5,
  colbacktitle=red!5,
  coltitle=black,
  fonttitle=\bfseries,
  bottomrule=0pt,
  toprule=0pt,
  leftrule=2pt,
  rightrule=2pt,
  titlerule=0pt,
  arc=0pt,
  outer arc=0pt,
  colframe=red,
}

\newtcolorbox{myboxnote}[1][]{
  breakable,
  title=#1,
  colback=orange!0,
  colbacktitle=orange!0,
  coltitle=black,
  fonttitle=\bfseries,
  bottomrule=0pt,
  toprule=0pt,
  leftrule=2pt,
  rightrule=2pt,
  titlerule=0pt,
  arc=0pt,
  outer arc=0pt,
  colframe=orange,
}

\newtcolorbox{myboxii}[1][]{
  breakable,
  freelance,
  title=#1,
  colback=white,
  colbacktitle=white,
  coltitle=black,
  fonttitle=\bfseries,
  bottomrule=0pt,
  boxrule=0pt,
  colframe=white,
  overlay unbroken and first={
  \draw[red!75!black,line width=3pt]
    ([xshift=5pt]frame.north west) -- 
    (frame.north west) -- 
    (frame.south west);
  \draw[red!75!black,line width=3pt]
    ([xshift=-5pt]frame.north east) -- 
    (frame.north east) -- 
    (frame.south east);
  },
  overlay unbroken app={
  \draw[red!75!black,line width=3pt,line cap=rect]
    (frame.south west) -- 
    ([xshift=5pt]frame.south west);
  \draw[red!75!black,line width=3pt,line cap=rect]
    (frame.south east) -- 
    ([xshift=-5pt]frame.south east);
  },
  overlay middle and last={
  \draw[red!75!black,line width=3pt]
    (frame.north west) -- 
    (frame.south west);
  \draw[red!75!black,line width=3pt]
    (frame.north east) -- 
    (frame.south east);
  },
  overlay last app={
  \draw[red!75!black,line width=3pt,line cap=rect]
    (frame.south west) --
    ([xshift=5pt]frame.south west);
  \draw[red!75!black,line width=3pt,line cap=rect]
    (frame.south east) --
    ([xshift=-5pt]frame.south east);
  },
}

\usepackage{fancyhdr} 
\usepackage{blindtext} 

\usepackage{forest}
\usepackage{xcolor}
\usepackage{tikz}
\usepackage{setspace}
\definecolor{lightblue}{RGB}{173,216,230}
\definecolor{lightgreen}{RGB}{144,238,144}
\definecolor{lightyellow}{RGB}{255,255,224}
\definecolor{lightpink}{RGB}{255,182,193}
\definecolor{hidden-draw}{RGB}{0,0,0}

\DeclareCaptionFont{black}{\color{black}}

\definecolor{myblue}{rgb}{0.9, 0.1, 0.94}
\definecolor{mygreen}{rgb}{0.64, 0.56, 0.88}
\definecolor{myyellow}{rgb}{0.68, 0.6, 0.1}
\definecolor{fancygreen}{rgb}{0.33, 0.68, 0.20}
\definecolor{salmon}{rgb}{0.94, 0.52, 0.49}
\definecolor{tablegreen}{rgb}{0.82, 0.94, 0.75}
\definecolor{tableblue}{rgb}{0.81, 0.90, 0.94}
\definecolor{tablered}{rgb}{0.97, 0.85, 0.85}
\definecolor{tableorange}{rgb}{0.96, 0.85, 0.81}

\newenvironment{itemize*}%
 {\leftmargini=10pt\begin{itemize}%
  \setlength{\itemsep}{0pt}%
  \setlength{\parskip}{0pt}%
  }%
 {\end{itemize}}
\newenvironment{enumerate*}%
 {\begin{enumerate}%
  \setlength{\itemsep}{0pt}%
  \setlength{\parskip}{0pt}}%
 {\end{enumerate}}

\usepackage{xcolor}
\usepackage{listings}

\newcommand\JSONnumbervaluestyle{\color{blue}}
\newcommand\JSONstringvaluestyle{\color{red}}

\newif\ifcolonfoundonthisline

\makeatletter

\lstdefinestyle{json}
{
  showstringspaces    = false,
  keywords            = {false,true},
  alsoletter          = 0123456789.,
  morestring          = [s]{"}{"},
  stringstyle         = \ifcolonfoundonthisline\JSONstringvaluestyle\fi,
  MoreSelectCharTable =%
    \lst@DefSaveDef{`:}\colon@json{\processColon@json},
  basicstyle          = \ttfamily,
  keywordstyle        = \ttfamily\bfseries,
}

\newcommand\processColon@json{%
  \colon@json%
  \ifnum\lst@mode=\lst@Pmode%
    \global\colonfoundonthislinetrue%
  \fi
}

\lst@AddToHook{Output}{%
  \ifcolonfoundonthisline%
    \ifnum\lst@mode=\lst@Pmode%
      \def\lst@thestyle{\JSONnumbervaluestyle}%
    \fi
  \fi
  \lsthk@DetectKeywords%
}

\lst@AddToHook{EOL}%
  {\global\colonfoundonthislinefalse}

\makeatother

\usepackage{etoolbox}
\usepackage{natbib}

\usepackage{url}
\newcounter{bibcount}
\makeatletter
\patchcmd{\@lbibitem}{\item[}{\item[\hfil\stepcounter{bibcount}{[\thebibcount]}}{}{}
\renewcommand\NAT@bibsetup%
  [1]{\setlength{\leftmargin}{\bibhang}\setlength{\itemindent}{-\parindent}%
      \setlength{\itemsep}{\bibsep}\setlength{\parsep}{\z@}}
\makeatother

\newtheorem{definition}{Definition}[section]

\newtheorem{mydef}{Definition}
\newtheorem*{definition*}{Definition}

\begin{document}

\title{Context Engineering 2.0: \\The Context of Context Engineering} 


\author{
    \centering
    Qishuo Hua\textsuperscript{1,3} \quad Lyumanshan Ye\textsuperscript{1,3} \quad Dayuan Fu\textsuperscript{2,3} \quad Yang Xiao\textsuperscript{2,3} \quad Xiaojie Cai\textsuperscript{1,3} \\[0ex] 
    Yunze Wu\textsuperscript{1,2,3} \quad Jifan Lin\textsuperscript{1,3} \quad Junfei Wang\textsuperscript{3} \quad Pengfei Liu\textsuperscript{1,2,3,\(\dagger\)}
}

\affil{\textsuperscript{1}SJTU \quad \textsuperscript{2}SII \quad \textsuperscript{3}GAIR}

\maketitle
\thispagestyle{fancy}
\fancyhead{}
\lhead{
  \raisebox{-0.3cm}{\includegraphics[height=0.95cm]{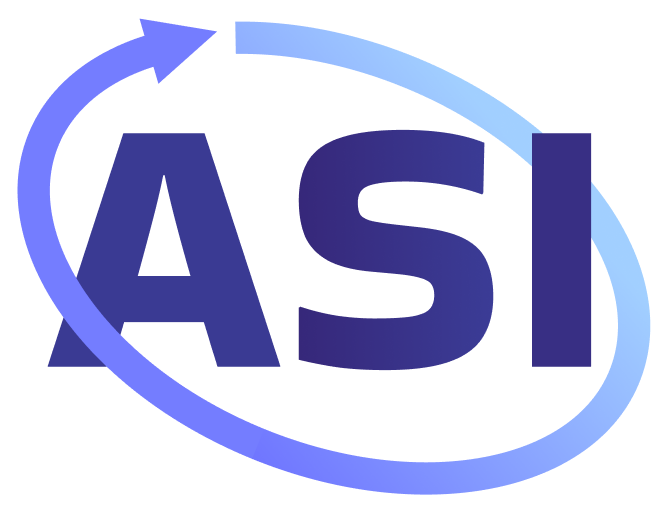}}
}
\rhead{%
  \raisebox{-0.2cm}{\includegraphics[height=0.7cm]{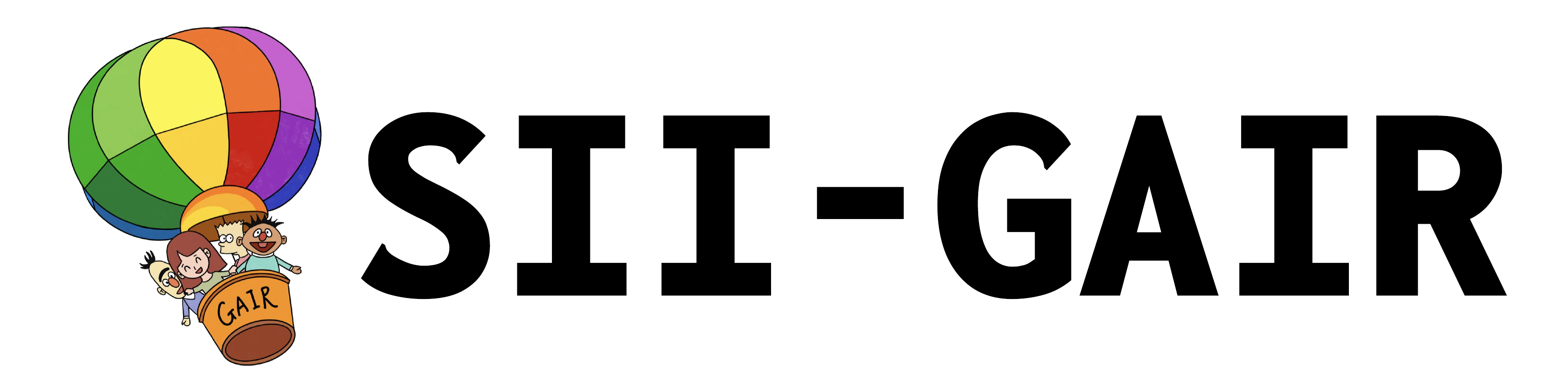}}%
}
\renewcommand{\headrulewidth}{0pt}
\setlength{\headsep}{2mm}
\footnotetext{† Corresponding author.}
\vspace{-20pt}

{\centering
\href{https://github.com/GAIR-NLP/Context-Engineering-2.0}{\textcolor{black}\faGithub\ Github} 
\quad 
\href{https://context.opensii.ai/}{\raisebox{-.15em}{\includegraphics[height=1em]{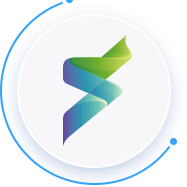}}\ SII Context}
\par}

\begin{tcolorbox}[colback=cyan!10, colframe=cyan!40, boxrule=0.5pt, arc=2mm]
\emph{``A person is the sum of their contexts.'' \hfill\textit{--- Authors}}
\end{tcolorbox}

\vspace{3pt}

\begin{abstract}

Karl Marx once wrote that ``the human essence is the ensemble of social relations''~\cite{marx1845theses}, suggesting that individuals are not isolated entities, but are fundamentally shaped by their interactions with other entities --- within which contexts play a constitutive and essential role. With the advent of computers and artificial intelligence, these contexts are no longer limited to purely human-human interactions: human-machine interactions are included as well. Then a central question emerges: \textbf{How can machines better understand our situations and purposes?} To address this challenge, researchers have recently ``developed'' the concept of \textit{context engineering}. Although it is often regarded as a recent innovation of the agent era, in fact, we argue that related practices can be traced back to over \textbf{20 years ago}. Since the early 1990s, it has evolved through distinct historical phases, each shaped by its intelligence level of machines: from early human-computer interaction (HCI) frameworks built around primitive computers, to today’s human-agent interaction (HAI) paradigms driven by intelligent agents, and potentially to human-level or even superhuman intelligence in the future. In this paper, we discuss the context of context engineering, provide a systematic definition, outline our perspective on its historical and conceptual landscape, and examine key design considerations for its practice. By addressing these questions, we aim to offer a conceptual foundation for context engineering and sketch its promising future. This paper serves as a stepping stone for a broader community effort toward systematic context engineering in AI systems.

\vspace{1.5em}

\begin{figure}[H]
    \centering
    \includegraphics[width=0.85\linewidth]{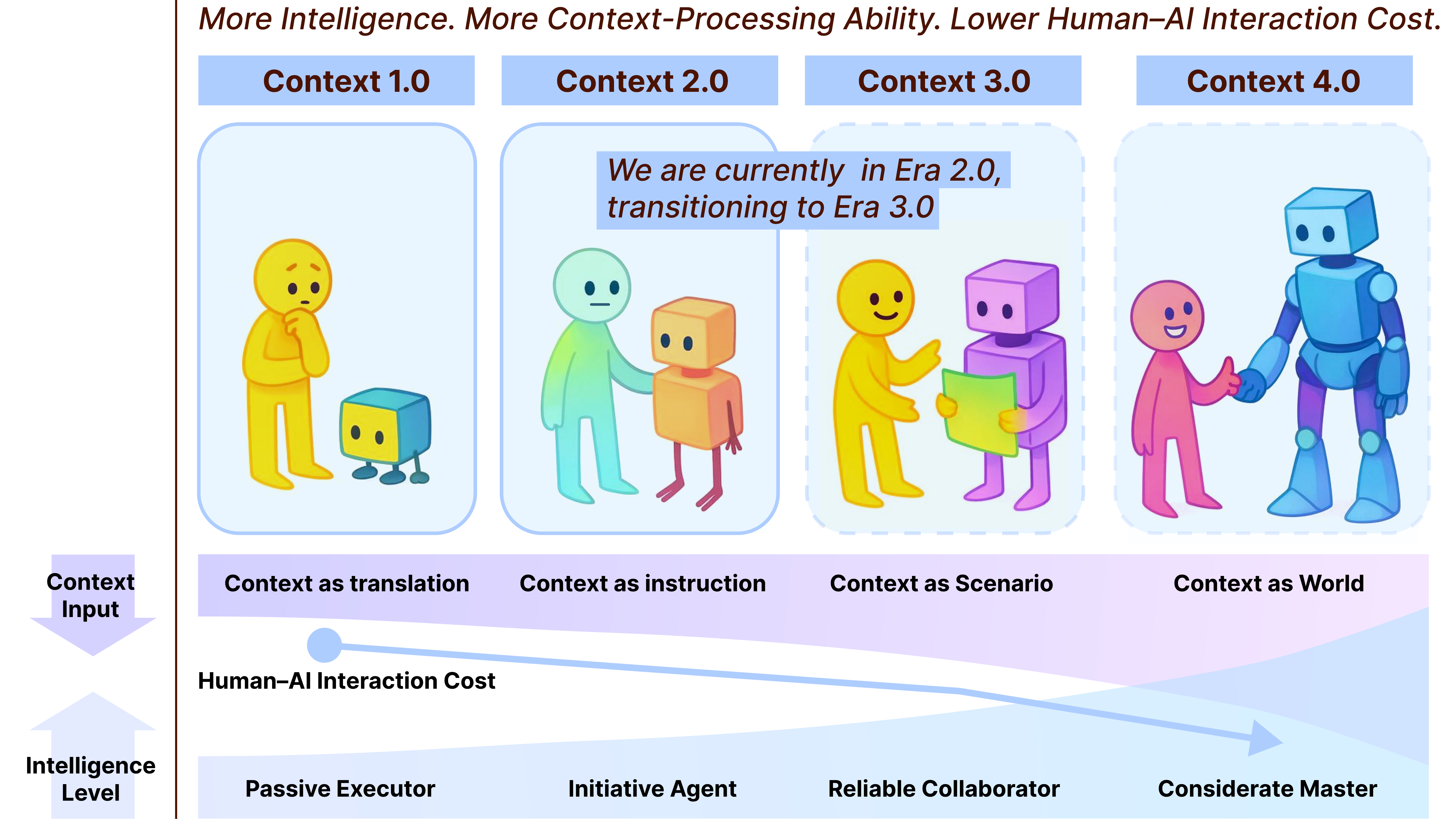}
    \caption{The Overview of context engineering 1.0 to context engineering 4.0, illustrating that more intelligence leads to greater context-processing ability and lower human-AI interaction cost.}
    \label{fig:placeholder}
\end{figure}
\end{abstract}









\newpage

\pagestyle{fancy}
\lhead{\rightmark}
\renewcommand{\headrulewidth}{0.7pt}
\setlength{\headsep}{5mm}

\clearpage

\onecolumn
\begin{spacing}{1.1}
\tableofcontents
\end{spacing}
\newpage

\section{Introduction}


In recent years, the rapid rise of large language models (LLM) and intelligent agents has drawn increasing attention to how context influences model behavior. Studies have shown that the content placed within the context window can significantly affect model performance~\cite{Prompt2021}. At the same time, there is growing demand for systems capable of multi-step reasoning and operating over long time horizons~\cite{yao2023reactsynergizingreasoningacting}. These trends make one question central: how can we enable machines to better understand and act upon human intent through effective context mechanisms, especially in long-horizon tasks?

To address this challenge, researchers have recently focused on \emph{context engineering}: the practice of designing, organizing, and managing contextual information so that machines can act in ways that align with human intentions~\cite{mei2025survey}. Recent years have witnessed extensive implementations of context engineering in LLMs and agents, including prompt engineering~\cite{Prompt2021, reynolds2021prompt, wei2022chain}, retrieval-augmented generation (RAG)~\cite{lewis2020retrieval, izacard2022distilling}, tool calling~\cite{yao2022react, schick2023toolformer}, and long-term memory mechanisms~\cite{wu2022memorizingtransformers, dai2019transformerxl}. These techniques expand a machine’s capacity to assimilate high-entropy context and have materially influenced the design of interactive systems.

Despite these advances, the field is often misunderstood. Context engineering is commonly perceived as a recent development, and ``context'' is often narrowly defined as dialogue history, system prompts, or agent-centric environmental input. In fact, context can be more broadly defined, and context engineering has been practiced \textbf{for more than 20 years}. Early research in ubiquitous computing, context-aware systems, and human–computer interaction established foundational principles and methods that remain relevant today~\cite{reeves2012envisioning, baldauf2007survey, preece1994human}. Recognizing this history is essential to understand both the current state of the field and its future potential. 

We argue that the development of context engineering should be viewed through a broader historical perspective rather than being confined to recent technical practices. By tracing its evolution over the past two decades, we can gain a deeper understanding of its underlying principles and recognize how different approaches to dealing with context have shaped the progress of intelligent systems. This view enables AI research to build upon its historical trajectory, establishing a robust and future-oriented foundation.


\begin{figure}[ht]
    \centering
    \includegraphics[width=0.9\linewidth]{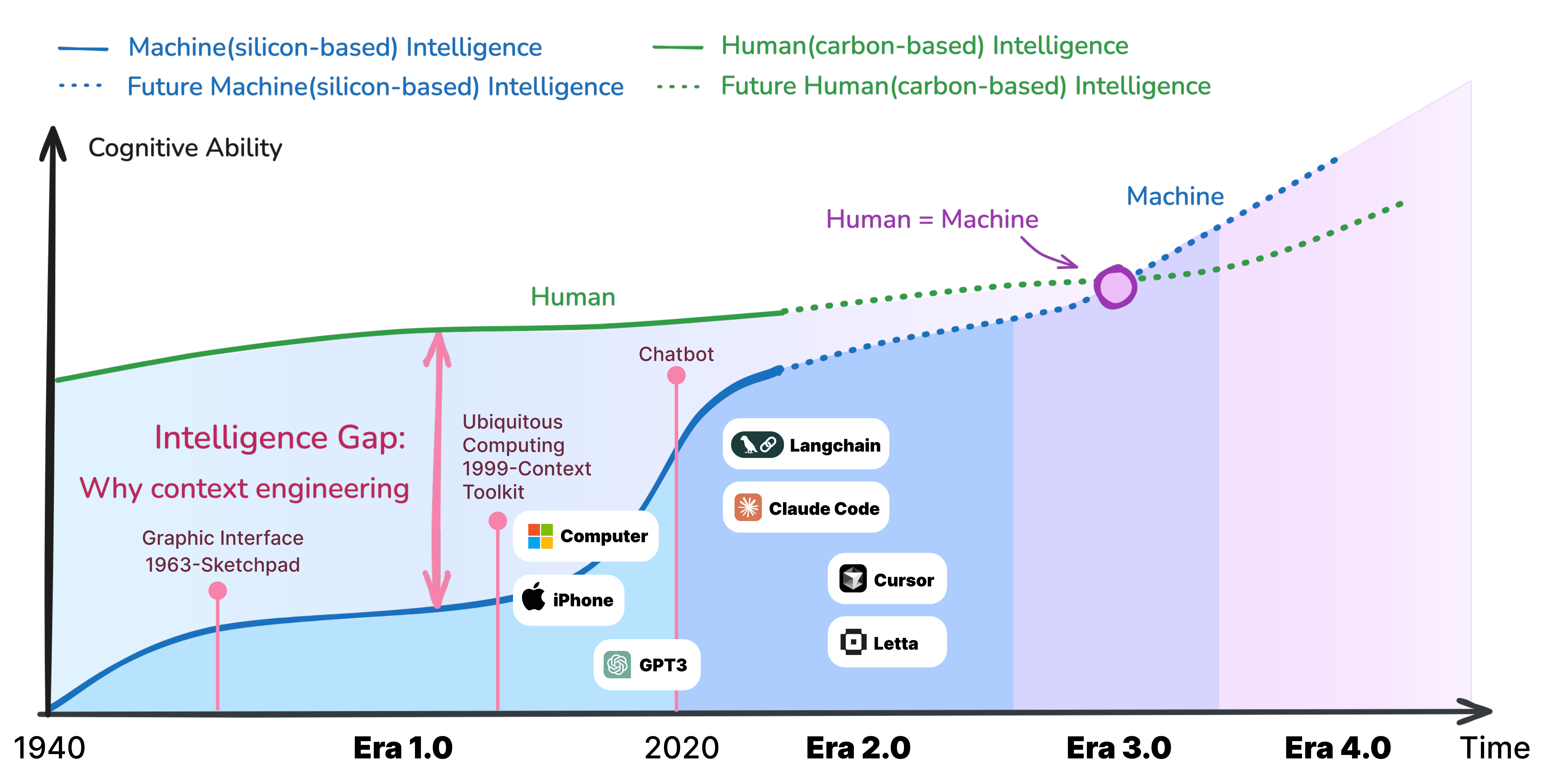}
    \caption{Trajectories of carbon-based and silicon-based cognitive abilities over time. The gap illustrates the motivation for context engineering.}
    \label{fig:curve_chart}
\end{figure}

From this broader perspective, context engineering can be viewed as a process of \emph{entropy reduction}. Machines, unlike humans, cannot effectively ``fill in the gaps'' during communication. When people interact, they rely on the listener’s ability to actively reduce information entropy --- the capacity to infer missing context through shared knowledge, emotional cues, and situational awareness~\cite{kapteijns2021sentence}. Machines, at least at present, lack this ability. As a result, we must ``preprocess'' contexts for them, compressing the original information into forms they can understand. This represents the core ``effort'' in context engineering: the effort you need to invest in transforming high-entropy contexts and intentions into low-entropy representations that machines can understand. As illustrated in Figure~\ref{fig:curve_chart}, context engineering has always existed to bridge the cognitive gap between human (carbon-based) and machine (silicon-based) intelligence. While carbon-based intelligence develops relatively slowly, silicon-based intelligence iterates at a much faster pace. Therefore, the key driver behind paradigm shifts lies in the rapid advancement of machine intelligence. The more intelligent machines are, the more natural context engineering becomes, and the lower the cost of human–machine interaction. 

\begin{figure}
    \centering
    \includegraphics[width=0.8\linewidth]{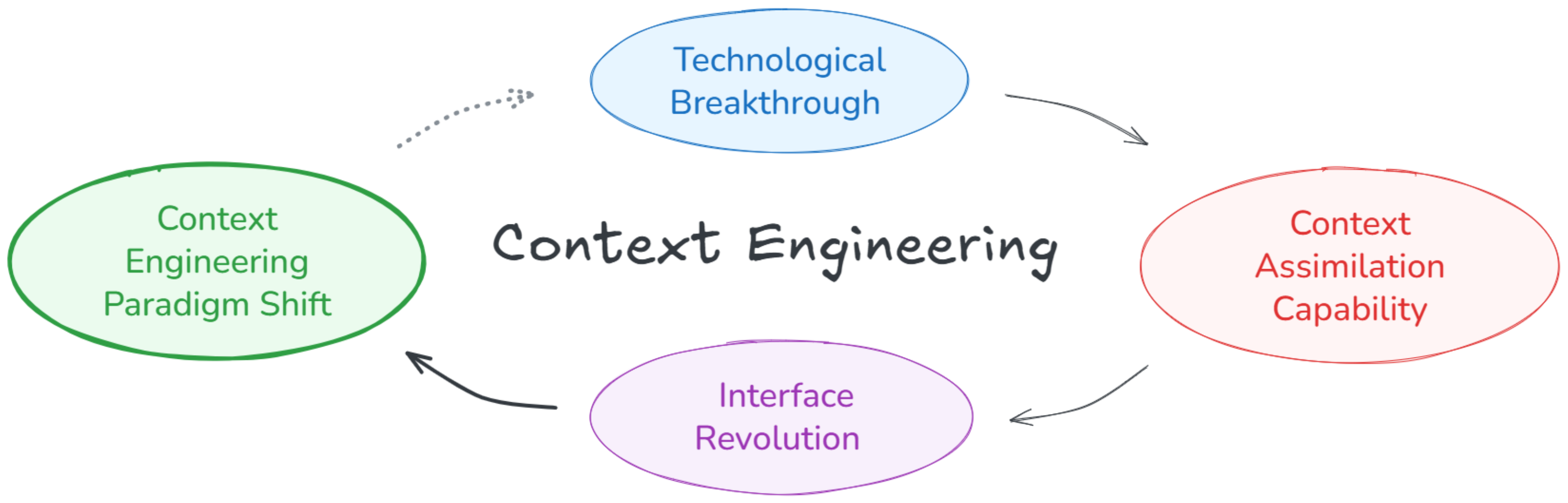}
    \caption{Evolutionary process in context engineering}
    \label{fig:evolutionv4}
\end{figure}

Each qualitative leap in machine intelligence triggers a fundamental revolution in human-machine interfaces. As illustrated in Figure~\ref{fig:evolutionv4}, \textbf{technological breakthroughs lead to a surge in context assimilation capability, which drives interface revolutions and ultimately results in paradigm shifts in context engineering}. These shifts are not gradual improvements, but a series of paradigm changes that fundamentally reshape how humans and machines communicate. Based on this recurring pattern, we can conceptualize the evolution of context engineering as a progression through \textbf{four distinct stages}: Context Engineering 1.0 --- primitive computing with structured, low-entropy inputs~\cite{Dey2001}; 2.0 --- intelligent agents capable of interpreting natural language and handling ambiguity~\cite{jennings1998roadmap}; 3.0 --- human-level intelligence, enabling nuanced communication and seamless collaboration~\cite{morris2023levels}; and 4.0 --- superhuman intelligence, where machines can proactively construct context and reveal needs that humans have not explicitly articulated. Each stage entails qualitatively different design trade-offs and highlights the shifting roles of humans and machines.



The core contributions of this paper are as follows. First, We situate context engineering within a broader historical perspective, tracing its origins before modern intelligent agents. Second, we present a systematic and broadly defined theoretical framework, including the essence of context engineering from an entropy reduction perspective and a four-stage evolutionary model reflecting its technical progress. Finally, we propose general design considerations by comparing typical practices, providing guidance for future intelligent system development.

In the remainder of this paper, we will first present our definition of context engineering and outline the theoretical framework that grounds our analysis. We will then trace the historical evolution of the field, with particular attention to the distinctive characteristics of the 1.0 and 2.0 eras. Building on this foundation, we discuss design considerations structured around three core dimensions: context collection, context management, and context usage. We analyze current practices, identify unique challenges, and explore emerging approaches that may shape future advancements.

\section{Theoretical Framework}

To better understand context, we begin by formalizing the fundamental constructs that define context in computational systems. Building on Dey’s foundational definition of context~\cite{dey2001understanding}, we express these constructs mathematically, providing a precise representation of context. Based on this formalization, we define context engineering and examine how it has evolved across different stages of machine intelligence.

\subsection{Formal Definition}

\begin{mydef}[Entity and Characterization]
Let $\mathcal{E}$ be the space of all entities (users, applications, objects, environments, etc.) and $\mathcal{F}$ be the space of all possible characterization information. Denote by $\mathcal{P}(\mathcal{F})$ the power set of $\mathcal{F}$. For any entity $e \in \mathcal{E}$, define the situational characterization function:
\begin{equation}
\text{Char}: \mathcal{E} \rightarrow \mathcal{P}(\mathcal{F})
\end{equation}
where $\text{Char}(e)$ returns the set of information characterizing entity $e$.
\end{mydef}




For example, when a user types ``Search related documentation for me'' in the Gemini CLI \footnote{\url{https://github.com/google-gemini/gemini-cli}}, the set of relevant entities can include the user, the Gemini CLI application, the terminal environment, external tools, memory modules, and backend model services. Here, $\text{Char}(e)$ can describe the user (e.g., the input prompt), the application (e.g., system instructions or configuration), the environment (e.g., current working directory), external tools (e.g., available plugins or search tools), short-term or long-term memory modules (e.g., session history or stored knowledge), and the model service (e.g., supported capabilities or response format).

\begin{mydef}[Interaction]
An interaction refers to any observable engagement between a user and an application, encompassing both explicit actions (e.g., clicks, commands) and implicit behaviors (e.g., attention patterns, environmental adjustments) that may influence or be influenced by the computational system.
\end{mydef}
In the Gemini CLI example, the explicit interaction is the user's command input, while implicit aspects can include terminal state, previously retrieved context, usage of memory modules, or tool invocation status.

\begin{mydef}[Context]
For a given user-application interaction, Context is defined as:
\begin{equation}
C = \bigcup_{e \in \mathcal{E}_{\text{rel}}} \text{Char}(e)
\end{equation}
where $\mathcal{E}_{\text{rel}} \subseteq \mathcal{E}$ is the set of entities considered relevant to the interaction.
\end{mydef}

This captures context as ``\textbf{any information that can be used to characterize the situation of entities that are considered relevant to the interaction between a user and an application}''~\cite{dey2001understanding}. 
In the same example, $\mathcal{E}_{\text{rel}}$ can include the user, the Gemini CLI application, the terminal environment, external tools, memory modules, and backend model services. The general context $C$ is then the aggregation of these characterizations.

\begin{mydef}[Context engineering]
Context engineering is the systematic process of designing and optimizing context collection, storage, management, and usage to enhance machine understanding and their task performance. Formally, it can be defined as:

\begin{equation}
\text{CE}: (C, \mathcal{T}) \rightarrow f_{\text{context}}
\end{equation}

where $C$ denotes the raw contextual information as defined in Equation (2), $\mathcal{T}$ represents the target task or application domain, and $f_{\text{context}}$ is the resulting context processing function that transforms and optimizes context representations for improved task performance.
$f_{\text{context}}$ encompasses a flexible composition of context processing operations:

\begin{equation}
f_{\text{context}}(C) = \mathcal{F}(\phi_1, \phi_2, \ldots, \phi_n)(C)
\end{equation}

where $\mathcal{F}$ represents a composition function that combines various context engineering operations ${\phi_i}$. 

\end{mydef}

The composition $\mathcal{F}$ can involve parallel processing, iterative refinement, conditional branching, or any combination of operations tailored to the specific application requirements. In practice, the operation set $\{\phi_i\}$ in context engineering 2.0 may include: (1) collecting relevant contextual information through sensors or other channels, (2) storing and managing it efficiently, (3) representing it in a consistent and interoperable format, (4) handling multimodal inputs from text, audio, or vision, (5) integrating and reusing past context (“self-baking”), (6) selecting the most relevant contextual elements, (7) sharing context across agents or systems, and (8) adapting context dynamically based on feedback or learned patterns.

\paragraph{The Scope of Context Engineering}

This definition deliberately avoids limiting context engineering to any specific technology or era. Whether the receiving entity is a 1990s primitive computer with a graphical interface or a 2025 agent, the fundamental challenge remains the same: how to make contexts and intentions accurately understood. The specific techniques and formats used in context engineering evolve with technology, but the core challenge of bridging the gap between human intentions and machine understanding remains constant.

\paragraph{Why This Definition Matters?}

This broader definition serves several important purposes. First, it connects current prompt engineering practices with the rich history of human-computer interface design, allowing us to learn from decades of accumulated knowledge. Second, it provides a theoretical framework that can explain why certain context designs work across different technologies and eras. Third, it offers a foundation for predicting how context engineering will evolve as machine understanding capabilities continue to advance. By understanding context engineering as a fundamental aspect of human-machine communication rather than a narrow technical practice, we can better appreciate both its historical development and its future trajectory. This perspective reveals context engineering not as a recent invention, but as an evolving discipline that will continue to adapt as the nature of human-machine interaction itself transforms.

\subsection{Stage Characterization}

To better understand context engineering, we must expand our view beyond the current moment and recognize it as a fundamental aspect of human-machine communication that has been continuously evolving for decades; this ongoing evolution can be characterized through four stages, each aligned with major advances in machine intelligence. Specifically, machine intelligence can be considered to progress through the following stages: (i) Primitive Computation, (ii) Agent-Centric Intelligence, (iii) Human-Level Intelligence, and (iv) Superhuman Intelligence.

\paragraph{Era 1.0: Primitive Computation (1990s-2020)}
In this stage, machines had only a very limited ability to interpret contexts. They could process structured inputs and recognize simple environmental cues, but lacked a deeper understanding of meaning or intent. Human–machine interactions relied on rigid, predefined formats, such as selecting from menus or using simple sensor data as input. Although this era went beyond binary-level commands, all contexts still had to be explicitly prepared and translated into formats the machine could directly process~\cite{shneiderman1987designing}.

\paragraph{Era 2.0: Agent-Centric Intelligence (2020–Present)}
The emergence of LLM, exemplified by the release of GPT-3 in 2020~\cite{floridi2020gpt, brown2020language}, marks a turning point for context engineering and agent-centric intelligence. Machines in this stage exhibit moderate intelligence, characterized by the ability to comprehend natural language inputs and infer some implicit intentions. Human–machine collaboration becomes increasingly viable, as users are able to express their needs conversationally and systems can interpret much of the underlying meaning. The context is no longer limited to explicitly defined signals; it can encompass ambiguity and incomplete information. Agents actively reason over contextual gaps, using advanced language understanding and in-context learning to provide more adaptive and responsive interactions~\cite{bommasani2021opportunities}.

\paragraph{Era 3.0: Human-Level Intelligence (Future)}
With anticipated breakthroughs, intelligent systems are expected to approach human-level reasoning and understanding~\cite{goertzel2021artificial}. In this stage, context engineering transcends the current pattern, enabling agents to sense contexts and assimilate high-entropy information like humans. The scope of interpretable context expands significantly, such as social cues, emotional states, and richer environmental dynamics. Such advances enable truly natural human–machine collaboration, with AI acting as knowledgeable and effective peers.

\paragraph{Era 4.0: Superhuman Intelligence (Speculative)}
As intelligent systems surpass human capabilities, they begin to possess a ``god’s eye view'', understanding human intentions more deeply than humans themselves. At this stage, the traditional subject–object relationship is inverted: instead of machines passively adapting to human-defined contexts, they actively construct new contexts for humans, uncover hidden needs, and guide human thinking. Signs of this transformation are already emerging, for example, in Go, professional players are learning novel, superhuman strategies from AI. In this way, machines become sources of insight and inspiration, fundamentally redefining the nature of human-machine collaboration~\cite{silver2016mastering}.

\section{Historical Evolution}

\begin{table}[ht]
\centering
\setlength\tabcolsep{2.5pt}
\renewcommand\arraystretch{1.1}
\footnotesize
\begin{tabular}{p{3cm}p{5.5cm}p{5.5cm}}
\toprule
\textbf{Aspect} & \textbf{Context Engineering 1.0} & \textbf{Context Engineering 2.0} \\
\midrule
Time Period & 1990s--2020 & 2020--Present \\
\midrule
Technical Background & Ubiquitous computing, context-aware systems, HCI & Large language models, agents, prompt engineering \\
Typical Systems & Context Toolkit, Cooltown, ContextPhone & \includegraphics[height=0.8em]{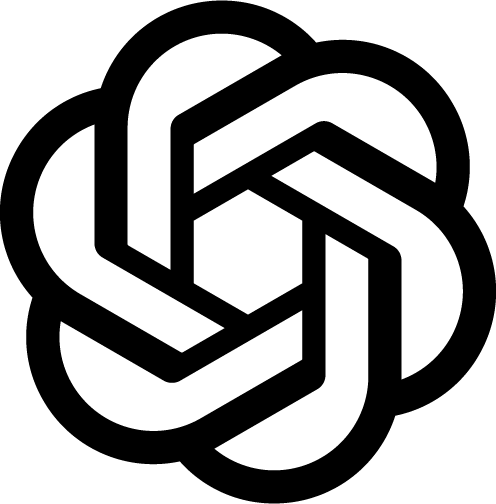} ChatGPT, \includegraphics[height=0.8em]{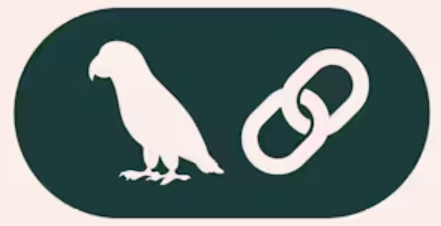} LangChain, \includegraphics[height=0.8em]{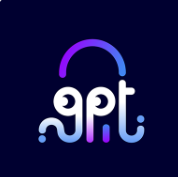} AutoGPT, \includegraphics[height=0.8em]{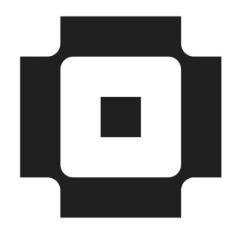} Letta \\
\midrule
Context Modalities & \includegraphics[height=0.8em]{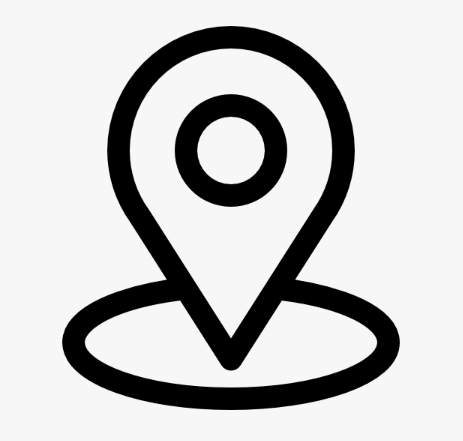} Location, \includegraphics[height=0.8em]{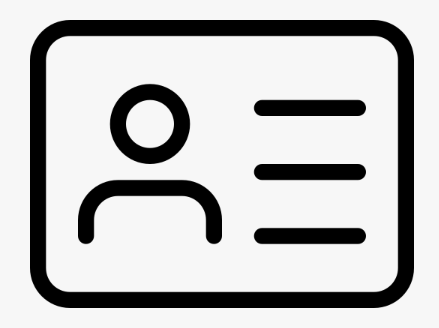} identity, activity, \includegraphics[height=0.8em]{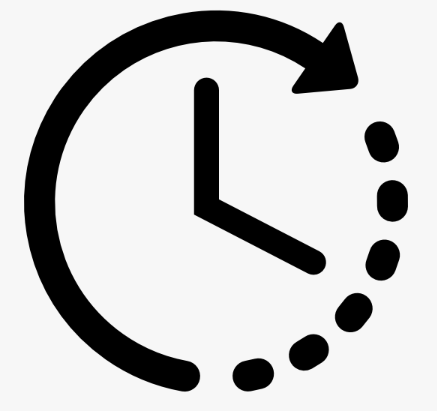} time, environment, device state & \includegraphics[height=0.8em]{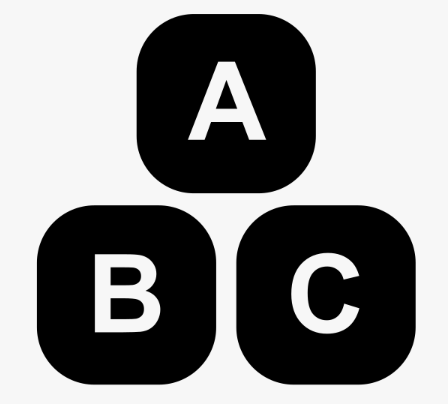} Token sequences, \includegraphics[height=0.8em]{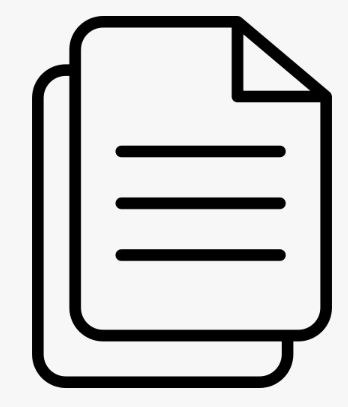} retrieved documents, \includegraphics[height=0.8em]{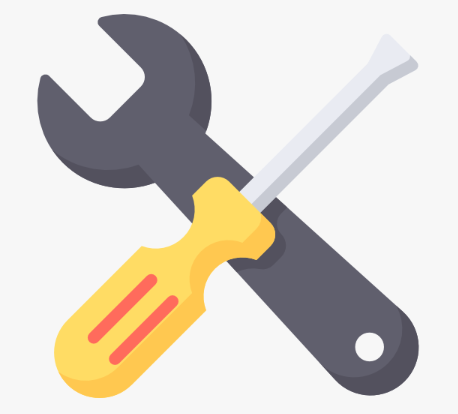} tool APIs, user history \\
Core Mechanisms & Sensor fusion, rule triggers & Prompting, RAG, CoT, memory agents \\
\midrule
Context Tolerance & Relatively Low & Relatively High \\
Human-likeness & Relatively Low & Relatively High \\
\bottomrule
\end{tabular}
\caption{Comparison between context engineering 1.0 and 2.0 across representative dimensions}
\label{tab:ce_1_2_comparison}
\end{table}

To better understand how context engineering has evolved, we compare its core characteristics across two major eras. The following sections examine the foundations of context engineering in its early development (Era 1.0) and the significant advancements that define the current landscape (Era 2.0).

\subsection{Over 20 Years Ago: Era 1.0}

We define context engineering as the practice of enabling effective understanding and communication between humans and machines. In the context engineering 1.0 era, humans are responsible for providing context, while primitive computers are the recipients of information. This era primarily spans the 1990s to 2020. In context engineering 1.0, humans struggle to adapt to machines: designers served as \emph{``intention translators''}, converting complex human intentions into structured, machine-readable formats. This stands in contrast to the modern era of AI with semantic understanding capabilities, where machines can autonomously interpret unstructured, high-entropy information.

\subsubsection{Technological Landscape}

To understand context engineering 1.0, we must first consider the technological landscape of the 1980s and 1990s. This period marked a transition from command-line interfaces (CLI) to graphical user interfaces (GUI), making computers more accessible to the general public. However, this shift did not eliminate the cognitive load on users. Instead, it redefined how humans interacted with machines, requiring precise adaptations to the limitations of the technologies of the time. In 1991, Mark Weiser introduced the concept of \emph{Ubiquitous Computing}~\cite{Weiser1991}, envisioning a seamless integration of computing into everyday environments. This vision suggested that devices could provide services without requiring active user input. Building on this idea, \emph{Context-Aware Computing} emerged as a framework to explore whether systems could sense user states, environments, and tasks to adapt their behavior dynamically~\cite{Schilit1994,AbowdDey1999}. This led to foundational questions: \emph{What exactly is ``context''? How can it be defined, processed, and made usable by machines?}

Despite these conceptual advances, the technological limitations of the time were significant. Machines could execute only pre-defined program logic and lacked the ability to understand natural language semantics, reason about problems, or handle errors effectively. This created a significant gap between human thinking and machine processing capabilities. To bridge this gap, \textbf{context engineering} provided strategies for making human intentions actionable by machines. Designers had to decompose complex goals into simple, structured components that machines could process. Clear interaction pathways needed to be designed to ensure machines could follow pre-determined logic. Additionally, feedback mechanisms were implemented to allow adjustments based on user input or environmental changes. These strategies were essential to overcoming the limitations of machines and enabling them to act on human intent effectively.

\subsubsection{Theoretical Foundations}
The early 2000s saw the emergence of robust theoretical frameworks for context engineering. Among them, Anind K. Dey’s 2001 definition of \emph{context} stood out as a cornerstone~\cite{Dey2001}:
\begin{quote}
    \textit{``Context is any information that can be used to characterize the situation of an entity. An entity is a person, place, or object that is considered relevant to the interaction between a user and an application, including the user and the applications themselves.''}
\end{quote}

This definition emphasized the multidimensional nature of contexts. Dey's work laid the foundation for systematically capturing and utilizing context to improve system adaptability and user experience. The theoretical work of this era was broad and deep~\cite{Baldauf2007}, emphasizing a holistic understanding of context. This contrasts with today's narrower focus, which often isolates specific aspects of context, such as chat history. The narrowing of this focus represents a step backward. Revisiting the foundational theories of context engineering 1.0 can help address this regression and inspire more comprehensive approaches to context-aware systems.

\subsubsection{Core Practices}

In the era of context engineering 1.0, a key innovation was the shift from traditional input devices, such as keyboards and mice, to a distributed, sensor-centric paradigm~\cite{Schilit1994,AbowdMynatt2000}. This transition reflected an emerging need to continuously capture richer contextual signals from both users and their surrounding environments. Building on this vision, Anind Dey introduced a general framework for context-aware systems that provided the conceptual and architectural foundation for this generation~\cite{Dey2001}. The framework was instantiated through the \emph{Context Toolkit}, which defined a modular and reusable framework to support the acquisition, interpretation, and delivery of context~\cite{salber1999context}. It was organized around five core abstractions: \emph{Context Widgets}, \emph{Interpreters}, \emph{Aggregators}, \emph{Services}, and \emph{Discoverers}. Widgets encapsulated sensors and exposed standardized interfaces; Interpreters derived higher-level meaning from raw contextual data; Aggregators integrated multiple context sources; Services offered application-level access to contextual functionality; and Discoverers enabled dynamic registration and discovery of context components~\cite{salber1999context}. This architectural separation of concerns established a practical foundation for scalable and adaptive context-aware systems, marking a decisive step in the formalization of early context engineering practices.

In sum, context engineering 1.0 laid the foundational thinking and architectural paradigms for building responsive systems. It brought together ideas from ubiquitous computing and HCI, established key abstractions through Anind Dey’s formalization~\cite{salber1999context}, and enabled practical implementations through modular, sensor-driven designs. Although limited in expressivity and scalability, this phase provided the essential groundwork for the more advanced designs that followed in context engineering 2.0 era.

\subsection{20 Years Later: Era 2.0}


The evolution from context engineering 1.0 to 2.0 represents a significant leap in machine intelligence, moving from primitive computing to intelligent agents~\cite{ye2025interactionintelligencedeepresearch}. Compared to era 1.0, which was dominated by rule-based reasoning and structured sensor input, agents introduce significant advancements throughout the context pipeline. These improvements span from how context is acquired to how raw signals are tolerated, interpreted, and ultimately used for intelligent behavior. In particular, the following shifts characterize the evolution:

\paragraph{Acquisition of Context: Advanced Sensors} 
In Era 2.0, sensors remain central to context acquisition. Advances in sensing technology, such as smartphones, wearables, and ambient devices, have greatly expanded the diversity and coverage of available contexts. Context engineering 2.0 emphasizes not only a broader diversity of sensors, but also the ability to extract diverse contextual signals from each individual sensor.

\begin{table}[ht]
\centering
\setlength\tabcolsep{2.5pt}
\renewcommand\arraystretch{1.1}
\footnotesize
\begin{tabular}{llll}
\toprule
\textbf{Category} & \textbf{Device/Collector} & \textbf{Collected Modalities} & \textbf{Example Input} \\
\midrule
    & Smartphone & Text, Image, Audio, Location, Touch & Messages, Photos, Voice \\
    & Computer (Laptop/PC) & Text, Image, Keystroke, Cursor & Mouse movement, Typing \\
\multirow{-3}{*}{Personal Computing}          & Smartwatch & Heart rate, Motion, Audio & Pulse, Steps \\
\midrule
& Smart glasses/AR headset & Video, Gaze, Voice, Scene context & Eye tracking, Ambient video \\
& VR/AR controller & Motion, Haptic feedback & Gesture, Button press \\
\multirow{-3}{*}{Immersive Technology}    & Smart speaker & Audio, Voice command & Conversations, Voice tone \\
\midrule
& Brain-computer interface & Neural signals, Emotion & EEG, Arousal, Cognitive load \\
& Skin sensors/wearables & Temperature, Galvanic response & Stress, Emotion, Touch pressure \\
\multirow{-3}{*}{Physiological Sensing}                   & Eye tracker & Gaze, Blink, Pupil dilation & Fixation patterns, Attention shift \\
\midrule
& Car system & Location, Gaze, Driving behavior & Driving style, Eye direction \\
& Home IoT devices & Environment, Sound, Motion & Temperature, Appliance use \\
\multirow{-3}{*}{Environmental Systems} & Online behavior tracking & Text, Clickstream, Scroll & Search intent, Interest patterns \\
\bottomrule
\end{tabular}
\caption{Representative Multimodal Context Collectors by Category}
\label{tab:context_collectors}
\end{table}

\paragraph{Tolerance for Raw Context: From Structured Inputs to Human-Native Signals} 
Before reaching human-level intelligence, a system's intelligence is principally governed by its degree of human-likeness, best measured by its tolerance for raw context, or in other words, the ability to consume and process high-entropy information input. In the 1.0 era of context-aware systems, input was limited to simple, structured signals like GPS coordinates, time of day, or predefined user states~\cite{Baldauf2007,Dey2001}. These features were easy to process, but required developers to define in advance what counts as meaningful context. In contrast, modern systems in the 2.0 era can interpret context from signals that resemble natural human expression, such as free-form text, images, or video~\cite{Radford2021,Alayrac2022}. This shift is not merely about adding new input types; it reflects a deeper improvement in the system’s ability to understand messy, ambiguous, and incomplete data. Enabled by advances in foundation models and multimodal perception, these systems can now handle inputs that were previously considered too raw or unstructured. As a result, context can be ingested directly in its native form, without the need for heavy pre-processing. This marks a fundamental step toward human-level flexibility in context interpretation.

\paragraph{Understanding and Utilization of Context: From Passive Sensing to Active Understanding and Collaboration} 
Context-aware systems in the 1.0 era typically operated under simple condition-action rules~\cite{Schilit1994,AbowdMynatt2000}. In other words, they sensed predefined signals and triggered fixed responses. For example, ``if location is office, then silence the phone.'' These systems respond according to where you are, but not to what you are doing. In contrast, 2.0 systems aim to actively interpret what the user is doing and collaborate to achieve shared goals. For example, when you are writing a research paper, the system can analyze your previous paragraphs and current writing intentions to suggest a suitable next section. It does not just sense your environment; it integrates into your workflow. That is what we call context-cooperative; we develop from \textbf{context-aware} to \textbf{context-cooperative} systems.


\section{Context Collection and Storage}





\definecolor{YellowFill}{HTML}{FFD88F}
\definecolor{YellowDraw}{HTML}{FFB03B}
\definecolor{PinkFill}{HTML}{FEBFC4}
\definecolor{PinkDraw}{HTML}{FF8A94}

\definecolor{BlueFill}{HTML}{9AC1FF}
\definecolor{BlueDraw}{HTML}{64A0FF}

\definecolor{PurpleFill}{HTML}{AB9CFF}
\definecolor{PurpleDraw}{HTML}{8374D4}

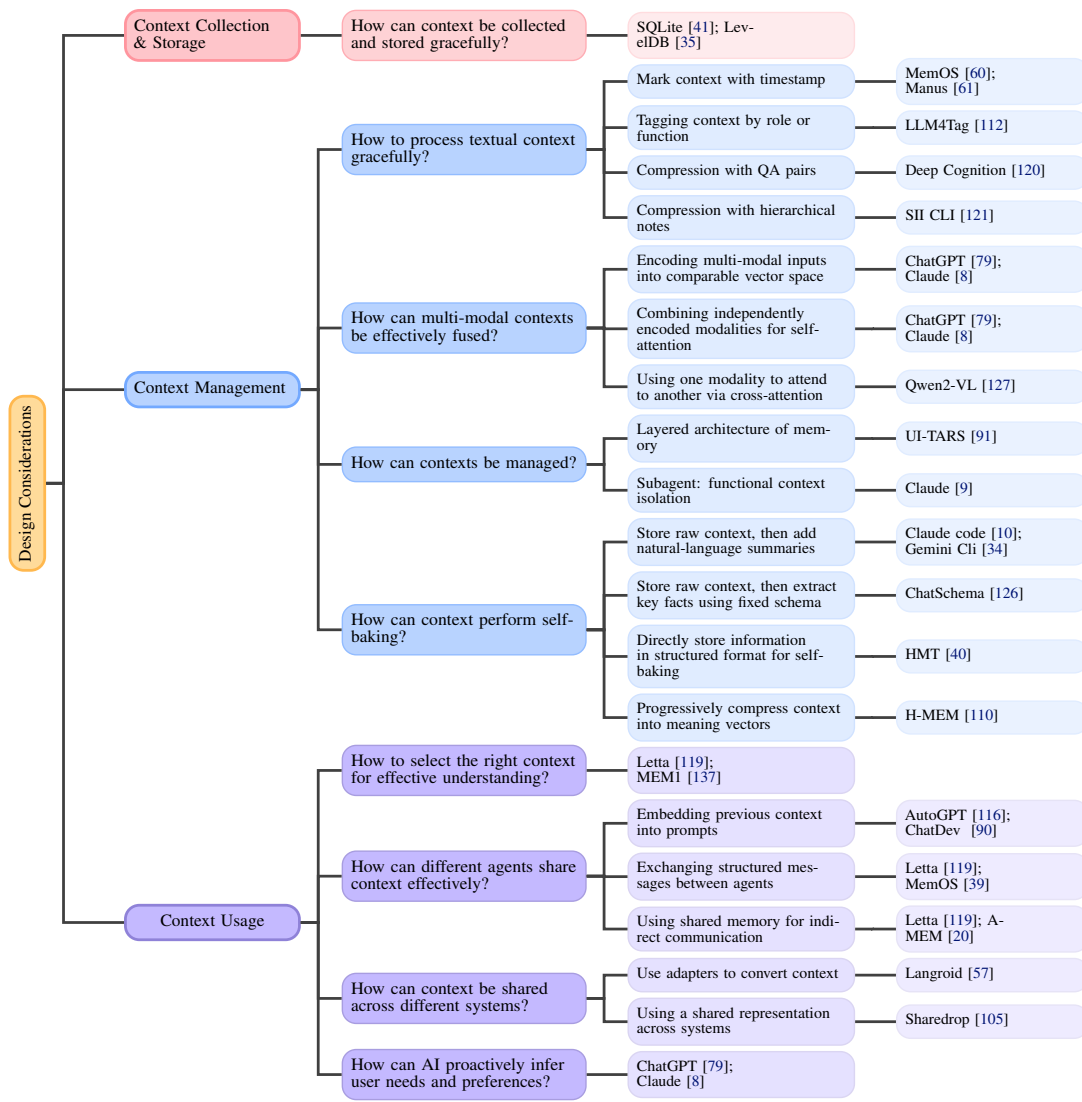
\begin{figure}[t!]
    \centering
    \tikzstyle{my-box}=[
        rectangle,
        rounded corners,
        text opacity=1,
        minimum height=0.2em,
        minimum width=0.2em,
        inner sep=2pt,
        align=left,
        fill opacity=.5,
    ]
    \tikzstyle{leaf}=[my-box]
    \resizebox{0.90\textwidth}{!}{%
        \begin{forest}
            forked edges,
            for tree={
                grow=east,
                reversed=true,
                anchor=base west,
                parent anchor=east,
                child anchor=west,
                base=left,
                font={\fontsize{5}{5}\selectfont},
                rectangle,
                draw=hidden-draw,
                rounded corners,
                align=left,
                minimum width=0.2em,
                edge+={darkgray, line width=0.8pt},
                s sep=1.5pt,
                inner xsep=2.5pt,
                inner ysep=2.5pt,
                ver/.style={rotate=90, child anchor=north, parent anchor=south, anchor=center},
            },
            where level=1{text width=4.5em,font={\fontsize{5}{5}\selectfont}}{},
            where level=2{text width=6.5em,font={\fontsize{5}{5}\selectfont}}{},
            where level=3{text width=6em,font={\fontsize{4.5}{4.5}\selectfont}}{},
            where level=4{text width=5em,font={\fontsize{4.5}{4.5}\selectfont}}{},
            where level=5{text width=9em,font={\fontsize{4.5}{4.5}\selectfont}}{},
            [
            Design Considerations, draw=YellowDraw, fill=YellowFill, draw opacity=0.9, fill opacity=0.9, thick, text=black, ver, 
                [\parbox{8em}{Context Collection \& Storage},
                    draw=PinkDraw, fill=PinkFill, draw opacity=0.9, fill opacity=0.9, thick, text=black,
                    where level=2{fill=PinkFill, fill opacity=0.7, draw=PinkDraw, draw opacity=0.7}{} ,
                    where level=3{fill=PinkFill, fill opacity=0.3, draw=PinkDraw, draw opacity=0.3}{} ,
                    where level>=4{fill=PinkFill, fill opacity=0.2, draw=PinkDraw, draw opacity=0.2}{} ,
                    [\parbox{13em}{How can context be collected and stored gracefully?}
                        [\parbox{9em}{SQLite~\citenumber{Hipp2000}; LevelDB~\citenumber{LevelDB2014}}]
                    ]
                ]
                [Context Management,
                    draw=BlueDraw, fill=BlueFill, draw opacity=0.9, fill opacity=0.7, thick, text=black,
                    where level=2{fill=BlueFill, fill opacity=0.7, draw=BlueDraw, draw opacity=0.7}{} ,
                    where level=3{fill=BlueFill, fill opacity=0.3, draw=BlueDraw, draw opacity=0.3}{} ,
                    where level>=4{fill=BlueFill, fill opacity=0.2, draw=BlueDraw, draw opacity=0.2}{} ,
                    [\parbox{13em}{How to process textual context gracefully?}
                        [\parbox{13em}{Mark context with timestamp}
                            [\parbox{9em}{MemOS~\citenumber{MemOS2025}; Manus~\citenumber{openmanus2025}}]
                        ]
                        [\parbox{13em}{Tagging context by role or function}
                            [\parbox{9em}{LLM4Tag~\citenumber{LLM4Tag2025}}]
                        ]
                        [\parbox{13em}{Compression with QA pairs}
                            [\parbox{9em}{Deep Cognition~\citenumber{opensii2025}}]
                        ]
                        [\parbox{13em}{Compression with hierarchical notes}
                            [\parbox{9em}{SII CLI~\citenumber{opensii2025cli}}]
                        ]
                    ]
                    [\parbox{13em}{How can multi-modal contexts be effectively fused?}
                        [\parbox{13em}{Encoding multi-modal inputs into comparable vector space}
                            [\parbox{9em}{ChatGPT~\citenumber{openai2024chatgpt}; Claude~\citenumber{anthropic2025claude}}]
                        ]
                        [\parbox{13em}{Combining independently encoded modalities for self-attention}
                            [\parbox{9em}{ChatGPT~\citenumber{openai2024chatgpt}; Claude~\citenumber{anthropic2025claude}}]
                        ]
                        [\parbox{13em}{Using one modality to attend to another via cross-attention}
                            [\parbox{9em}{Qwen2-VL~\citenumber{wang2024qwen2vlenhancingvisionlanguagemodels}}]
                        ]
                    ]
                    [\parbox{13em}{How can contexts be managed?}
                        [\parbox{13em}{Layered architecture of memory}
                            [\parbox{9em}{UI-TARS~\citenumber{qin2025uitars}}]
                        ]
                        [\parbox{13em}{Subagent: functional context isolation}
                            [\parbox{9em}{Claude~\citenumber{anthropic2025subagents}}]
                        ]
                    ]
                    [\parbox{13em}{How can context perform self-baking?}
                        [\parbox{13em}{Store raw context, then add natural-language summaries}
                            [\parbox{9em}{Claude code~\citenumber{claudecode2025};
                            Gemini Cli~\citenumber{geminicli2025}}]
                        ]
                        [\parbox{13em}{Store raw context, then extract key facts using fixed schema}
                            [\parbox{9em}{ChatSchema~\citenumber{wang2024chatschemapipelineextractingstructured}}]
                        ]
                        [\parbox{13em}{Directly store information in structured format for self-baking}
                            [\parbox{9em}{HMT~\citenumber{he2025hmthierarchicalmemorytransformer}}]
                        ]
                        [\parbox{13em}{Progressively compress context into meaning vectors}
                            [\parbox{9em}{H-MEM~\citenumber{H_MEM2025}}]
                        ]
                    ]
                ]
                [\parbox{9em}{\centering Context Usage},
                    draw=PurpleDraw, fill=PurpleFill, draw opacity=0.9, fill opacity=0.7, thick, text=black,
                    where level=2{fill=PurpleFill, fill opacity=0.7, draw=PurpleDraw, draw opacity=0.7}{} ,
                    where level=3{fill=PurpleFill, fill opacity=0.3, draw=PurpleDraw, draw opacity=0.3}{} ,
                    where level>=4{fill=PurpleFill, fill opacity=0.2, draw=PurpleDraw, draw opacity=0.2}{} ,
                    [\parbox{13em}{How to select the right context for effective understanding?}
                        [\parbox{9em}{Letta~\citenumber{letta}; MEM1~\citenumber{mem1}}]
                    ]
                    [\parbox{13em}{How can different agents share context effectively?}
                        [\parbox{13em}{Embedding previous context into prompts}
                            [\parbox{9em}{AutoGPT~\citenumber{Team2025AutoGPT}; ChatDev ~\citenumber{qian2024chatdevcommunicativeagentssoftware}}]
                        ]
                        [\parbox{13em}{Exchanging structured messages between agents}
                            [\parbox{9em}{Letta~\citenumber{letta}; MemOS~\citenumber{memos}}]
                        ]
                        [\parbox{13em}{Using shared memory for indirect communication}
                            [\parbox{9em}{Letta~\citenumber{letta}; A-MEM~\citenumber{amem}}]
                        ]
                    ]
                    [\parbox{13em}{How can context be shared across different systems?}
                        [\parbox{13em}{Use adapters to convert context}
                            [\parbox{9em}{Langroid~\citenumber{langroid2025}}]
                        ]
                        [\parbox{13em}{Using a shared representation across systems}
                            [\parbox{9em}{Sharedrop~\citenumber{sharedrop2025}}]
                        ]
                    ]
                    [\parbox{13em}{How can AI proactively infer user needs and preferences?}
                        [\parbox{9em}{ChatGPT~\citenumber{openai2024chatgpt}; Claude~\citenumber{anthropic2025claude}}]
                    ]
                ]
            ]
        \end{forest}
   }
    \caption{Design considerations of context engineering across different eras, note that the examples listed only cover a subset}
    \label{fig:context_taxonomy}
\end{figure}


As computing environments become more complex, there is a growing need for enhanced context collection capabilities. In response, advances in sensing technologies and AI now make it possible to collect context from richer sources (or sensors). Storage has also diversified across local devices, network servers, cloud platforms, etc., each with different trade-offs in latency, capacity, scalability, and security. 
Two fundamental design principles guide this process. The Minimal Sufficiency Principle states that systems should collect and store only the information necessary to support a task: The context value lies in sufficiency, not volume. The Semantic Continuity Principle emphasizes that the purpose of context is to maintain continuity of meaning, rather than merely continuity of data. Together, these principles shape how context should be collected and preserved in intelligent and reliable systems.

\subsection{Typical Strategies in Era 1.0 and 2.0}

In the early stages, context was collected primarily on a single device, such as a desktop computer or an early smartphone, using limited sensors (GPS, clock, keyboard/mouse events) or application logs that recorded usage patterns and user interactions~\cite{Schilit1994,Dey2001,Abowd2002}.
Storage practices in this period were largely local. Context data was typically recorded as log files or structured documents within the local file system or stored in simple local databases. Temporary states, such as recent user input or window activities, were often kept in memory caches or temporary folders and discarded when the system was shut down. Although some systems attempted to upload context to centralized servers, these efforts were constrained by high latency and unstable network connectivity~\cite{Satyanarayanan2001}. In general, storage strategies in this era prioritized standalone availability on a single device rather than cross-device synchronization or secure data protection.

With technological progress, context collection became distributed across multiple endpoints, including smartphones, wearables, home sensors, cloud services, and third-party APIs~\cite{Lane2010,Miluzzo2008}. Agents integrated multimodal signals into continuous context streams~\cite{baltrusaitis2019multimodal}. 
Storage practices of this period commonly adopted a layered architecture, with storage strategies determined by the intended usage of the data. For example, short-lived or frequently accessed data may be cached in fast-access memory or at edge nodes to minimize latency. Data requiring medium-term retention, such as activity records or user preferences, can be stored in local embedded databases (e.g. SQLite~\cite{Hipp2000}, LevelDB~\cite{LevelDB2014}, RocksDB~\cite{RocksDB2013}) or, where security is paramount, within OS-backed secure storage or hardware security modules~\cite{Kostiainen2012}. For long-term persistence, scalability, and cross-device synchronization, cloud storage platforms and remote server databases may be used.

In the case of code agents, many tasks may run over long periods of time and often span multiple sessions, making it impractical to rely solely on the context window, which is both short-term and limited in capacity. To address this, systems periodically store the task state and progress in long-term memory so that an agent can resume work after interruption by restoring the relevant context. Such long-term memory can be maintained in local databases or secure storage, supported by cloud or remote services for cross-device synchronization, and, in some cases, even embedded into model parameters to provide more stable continuity and long-term adaptability. For example, Claude Code demonstrates a practical approach by maintaining structured notes, where key information is periodically written out of the context window into external memory and retrieved when needed. This strategy provides a lightweight but persistent form of memory, allowing the agent to track progress and avoid loss of information in complex tasks. More generally, this mechanism has been shown to support long-horizon activities, such as managing strategies over thousands of steps in a Pokémon game and then resuming seamlessly after resets~\cite{Anthropic2025Effective}. In this way, structured external memory extends the agent’s planning horizon far beyond the limits of a short and transient context window.

\subsection{Human-Level Context Ecosystem}

In the 3.0 era, AI systems achieve contextual awareness that rivals human perception. They may seamlessly collect tactile information (e.g. texture, pressure, temperature), recreating the sensory experience of human touch. Through smell and taste, they interpret environmental conditions, detecting smoke as a danger signal or assessing food freshness. In addition, they capture intent and emotion from vocal tone, pauses, eye contact, and even silence, understanding the subtle social contexts that define human interaction.
Systems with human-level intelligence unify the context into a long-term \emph{personal digital memory}. Storage is no longer just about preserving data, but serves as a dynamic cognitive infrastructure: capable of autonomously organizing, refining context to support continuous reasoning and interaction across scenarios and over time, as well as achieving human-like ``forgetting'' and ``recalling'' capabilities. Data flows securely between local and cloud environments, ensuring that users retain absolute control over sensitive information while still benefiting from global knowledge and computational resources, enabling natural human–machine symbiosis.


\section{Context Management}


\subsection{Textual Context Processing}


Effective context engineering isn't just about collecting raw context, it is also about how we process it. A well-designed processing approach forms the foundation for everything that follows: interpretation, compression, and retrieval. It allows systems to focus on what matters, learn from past experiences, and build sustained long-term understanding. This section examines the fundamental question: How do we process raw textual context to get the best results? Beyond simply storing raw context, such as multi-turn dialogues, we examine several commonly used designs, along with their respective trade-offs.




\paragraph{Mark Context with Timestamp} A common design is to attach a timestamp to each piece of information, preserving the order in which it was generated. This method is popular in chatbots and user activity monitoring due to its simplicity and low maintenance cost. However, this approach suffers from several limitations. Although timestamps preserve temporal order, they provide no semantic structure, making it difficult to capture long-range dependencies or retrieve relevant information efficiently. As interactions accumulate, the sequence grows linearly, leading to scalability issues in both storage and reasoning~\cite{MemOS2025,LangChainBlog2025}.

\paragraph{Tagging Context by Functional and Semantic Attributes} This approach organizes contextual information by explicitly tagging each entry with a functional role, such as ``goal'', ``decision'', and ``action'', to make each entry easier to interpret. Recent systems enable tagging from multiple dimensions, including priority levels, source information, etc., thereby supporting more efficient retrieval and context management~\cite{CAIM2025,LLM4Tag2025}. Although this helps clarify the meaning of each piece of information, it can be slightly rigid and may limit more flexible reasoning or creative synthesis.



\paragraph{Compression with QA Pairs} This method reformulates context into distinct question–answer pairs to improve retrieval efficiency, particularly in applications such as search engines or FAQ-based systems~\cite{LlamaIndex2024}. However, it disrupts the original flow of ideas, making it less suitable for tasks that require a comprehensive understanding of the context, such as summarizing or reasoning~\cite{EXIT2024,ContextAwareCompression2025}.

\paragraph{Compression with Hierarchical Notes} This method organizes information in a tree-like structure, where broad concepts branch into increasingly specific sub-points. Although this structure helps to present ideas clearly, it primarily reflects how information is grouped rather than how ideas are logically connected. Relationships such as cause and effect or evidence and conclusion are often not represented. Furthermore, this design does not capture how understanding evolves over time, for example, when new insights emerge or existing ideas are revised~\cite{opensii2025, LIFT2025,PromptCompression2024, Che2024Hierarchical}.

\vspace{0.5em}



\subsection{Multi-Modal Context Processing}

\begin{figure}[ht]
    \centering
\includegraphics[width=1\linewidth]{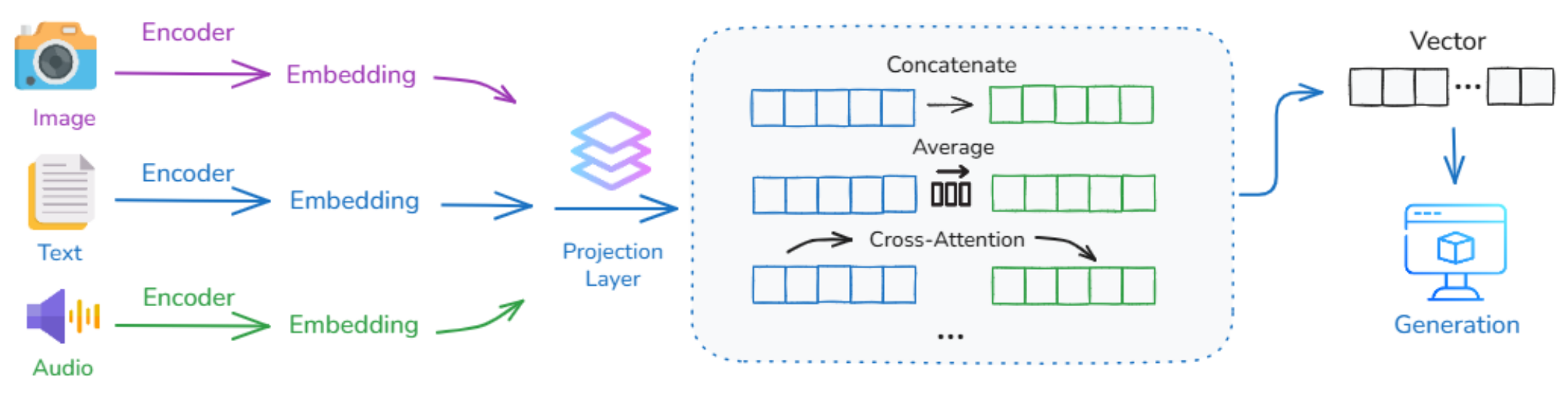}
    \caption{An example workflow for processing multimodal context with hybrid strategies}
    \label{fig:multimodal}
\end{figure}


Context in LLM-based systems is becoming increasingly multimodal, including text, images, audio, video, code, sensor data, and even environmental states. As machines interact with real or simulated environments, they must be able to integrate information across modalities into a coherent and unified representation rather than processing each modality in isolation. A core challenge lies in the heterogeneity of these modalities: they differ in structure, information density, and temporal dynamics. For instance, text is discrete and sequential; images are high-dimensional and spatial; audio is continuous and unfolds over time. How can these be jointly encoded, compared, and reasoned over? In era 2.0, we identify several common strategies:




\paragraph{Mapping Multimodal Inputs into a Comparable Vector Space} This method transforms inputs from different modalities, such as text, images, and video, into a shared vector space, so that their meanings can be directly compared. Each modality is first processed by its own encoder.
Since these vectors initially live in separate representation spaces with different statistical properties, each is then passed through a learned projection layer that maps it into a shared embedding space of fixed dimensionality. In this space, semantically related content from different modalities is positioned close together, while unrelated content is pushed apart~\cite{kosmos2,perceiver,kosmos2.5}. 
    
\paragraph{Combining Different Modalities for Self-Attention} 
After being projected into a shared embedding space, modality-specific tokens are jointly processed by a single Transformer architecture. In this unified self-attention mechanism, text and visual tokens attend to each other at every layer, allowing fine-grained cross-modal alignment and reasoning.
This approach, adopted by modern multimodal LLMs, enables the model to capture detailed correspondences, such as which phrase refers to which region of an image, rather than relying on the shallow concatenation of independent embeddings~\cite{openai2024chatgpt,anthropic2025claude}.

\paragraph{Using One Modality to Attend to Another via Cross-Attention} This method uses cross-attention layers to allow one modality (such as text) to directly focus on specific parts of another modality (such as images). Concretely, features of one modality are used as queries, while features of the other modality are treated as keys and values in the attention mechanism~\cite{vaswani2017attention}. This setup enables the model to retrieve relevant information across modalities in a targeted and flexible way. Cross-attention mechanisms can be flexibly implemented either as dedicated modules before the main Transformer architecture or embedded within the Transformer blocks themselves, depending on the overall system design~\cite{racm3, flamingo}. However, traditional designs typically require specifying which modalities interact, whereas the human brain can flexibly integrate information across sensory and memory channels without relying on such fixed mappings.

\subsection{Context Organization}

\subsubsection{Layered Architecture of Memory}

Managing information effectively across different time scales is a fundamental challenge in AI systems. As Andrej Karpathy puts it, LLMs can be viewed through an operating system analogy: the model acts like a CPU, while its context window resembles RAM--- fast but capacity-limited working memory. Just as an Operating System decides what data to load into RAM, context engineering determines what information should enter the window for effective reasoning~\cite{langchain2025context}. Like what we did in system, AI architectures benefit from separating memory into distinct layers based on temporal relevance and importance. This hierarchical approach allows systems to maintain quick access to recently relevant information while preserving valuable knowledge in more stable, long-term storage. For example, LeadResearcher systems store research plans in persistent memory when handling ultra-long contexts ($>$200k tokens), preventing key information from being lost due to context window limits~\cite{claude2025build}. The core insight is that different types of information require different retention strategies. Recent contexts need fast retrieval, but may become irrelevant quickly, while important patterns and learned knowledge should persist across sessions. By organizing the memory hierarchically, systems can optimize both responsiveness and storage efficiency.

The framework we present here focuses on a two-layer model for clarity, but the principles extend naturally to more complex architectures. Real implementations often include additional intermediate layers, such as working memory for active processing, episodic buffers for recent events, or specialized caches for different domains, each with its own temporal characteristics and selection criteria~\cite{qin2025uitars, stackademic2025layered}.

\begin{definition}[Short-term Memory]
Short-term memory is defined as the subset of context with high temporal relevance, selected by a processing function:
\begin{equation}
M_s = f_{\text{short}} \left({c \in C : w_{\text{temporal}}(c) > \theta_s}\right)
\end{equation}
where $w_{\text{temporal}}(c)$ is the temporal relevance weight function of context element $c$, $\theta_s$ is the temporal relevance threshold for short-term memory, and $f_{\text{short}}$ is a processing function that may involve human judgment, heuristic filtering, or system-level operations.
\end{definition}

\begin{definition}[Long-term Memory]
Long-term memory is defined as the processed and abstracted subset of context with high importance:
\begin{equation}
M_l = f_{\text{long}} \left({c \in C : w_{\text{importance}}(c) > \theta_l \land w_{\text{temporal}}(c) \leq \theta_s}\right)
\end{equation}
where $w_{\text{importance}}(c)$ is the importance weight of context element $c$, $\theta_l$ is the importance threshold for long-term memory, and $f_{\text{long}}$ is a composite function that may combine selection, abstraction, and compression to generate stable representations.
\end{definition}

\begin{definition}[Memory Transfer]
The transfer from short-term to long-term memory is defined as:
\begin{equation}
f_\text{transfer}: M_s \rightarrow M_l
\end{equation}
This transfer function represents the consolidation process where frequently accessed or highly important information in short-term memory is processed to become part of long-term memory. The transfer is governed by factors such as repetition frequency, emotional significance, and relevance to existing knowledge structures.
\end{definition}

\subsubsection{Context Isolation}

\paragraph{Subagent} 
A subagent provides an alternative way around context limitations while also reducing the risk of context pollution, representing an emerging strategy for effective context management through functional context isolation~\cite{Anthropic2025Effective}. The Claude Code subagent system illustrates this principle: Each subagent is a specialized AI assistant with its own isolated context window, custom system prompt, and restricted tool permissions. When a task matches the expertise of a subagent, the main system can delegate it to that unit, which then operates independently without contaminating the primary context of conversation. 
This principle can be applied to context selection by implementing separation along functional dimensions (e.g., analysis, execution, validation) or hierarchical layers (e.g., planning, implementation, review), where each isolated unit receives only the minimum permissions required for its specific responsibilities, improving both system reliability and interpretability~\cite{anthropic2025subagents}. Unlike static retrieval approaches such as RAG, the LeadResearcher first makes a plan. If the available information is insufficient, it can issue further searches, adjust keywords, or create new sub-tasks. The subagents may work in parallel, while the LeadResearcher summarizes interim results and decides on the next steps. This feedback loop helps the system converge on high-quality answers~\cite{claude2025build}.

\paragraph{Lightweight References} 
Context isolation often relies on storing large information externally and exposing only lightweight references in the model’s window. In the sandbox approach, as used in HuggingFace’s CodeAgent, bulky output is stored in a separate sandbox and retrieved only when needed. In this way, the model interacts only with concise references, while the sandbox holds the full data and provides it on demand. A similar principle applies to schema-based state objects, where heavy elements such as files or logs remain in external storage and only selected fields are surfaced. Both approaches reduce token overhead while retaining access to the complete context when required~\cite{Lance2025Context}.


\subsection{Context Abstraction}

\begin{figure}[ht]
    \centering
\includegraphics[width=1\linewidth]{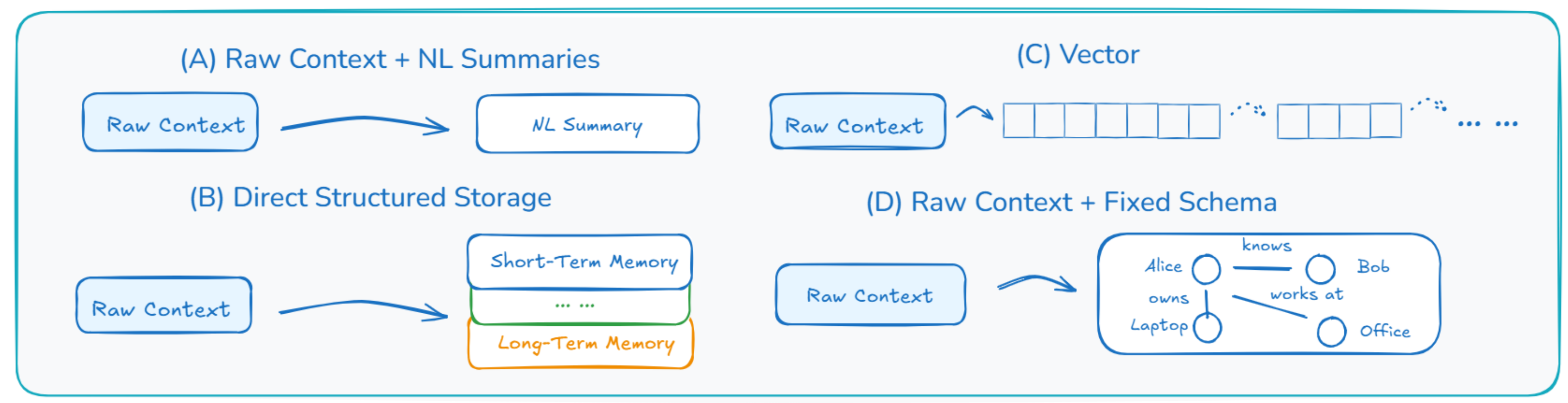}
    \caption{Representative designs for self-baking in Era 2.0}
    \label{fig:bakingv3}
\end{figure}

Raw context, such as dialogue turns, tool output, and retrieved documents, can accumulate quickly. If left unprocessed, the growing history can quickly overwhelm the system, making it difficult to identify what is truly important for future reasoning or decision making. To manage this, a crucial capability for scalable agents is \textbf{context abstraction}: converting raw context into more compact, structured representations. We refer to this process as \textit{self-baking}: the agent selectively \textbf{digests} its own context into persistent knowledge structures. This mirrors human cognitive processes, where episodic memories give rise to semantic memory, or where repeated actions are abstracted into habits. Ultimately, self-baking is what \textbf{separates memory \textit{storage} from \textit{learning}}. Without self-baking, agents simply recall; with it, they accumulate knowledge.




\paragraph{Using Hierarchical Memory Architectures} 

Current systems typically adopt an architectural principle: hierarchical memory organization. Hierarchical memory provides a principled way to manage growing contexts by organizing information at different levels of abstraction. At the base layer, raw contexts are stored to ensure that fine-grained details remain accessible. As the volume of context increases, these raw items are progressively summarized into more abstract representations, which are then passed to the next layer. New information typically enters at the lowest layer and is gradually ``baked'' upward, allowing the system to scale without overwhelming the context window while still retaining retrievable links back to the original details. In this sense, hierarchical memory complements the distinction between short-term and long-term memory: the raw context typically resides in lower short-term memory, while more abstract representations tend to correspond to long-term memory~\cite{he2025hmthierarchicalmemorytransformer,LayeredMemory2025}.



\paragraph{Add Natural-Language Summaries} In this pattern, the system stores the full context in its original, unstructured form. In addition, it periodically generates summaries that provide a compressed view of what has happened so far. These summaries, usually written in natural language, give a quick overview of recent events and can be created either manually or automatically. For example, a chatbot might store the full dialogue history and generate a brief paragraph that summarizes the recent conversation. As the number of summaries grows, the systems may apply multilevel summarization (summarizing older summaries into higher-level overviews) or use specific strategies to drop less useful ones based on time or importance~\cite{Lance2025Context}. These summaries help the system focus on key information while still keeping the full details available if needed. This method is simple and flexible. However, because the summaries are just plain text, they lack structure. This makes it difficult for the system to understand the connections between events or perform deeper reasoning over the context~\cite{DailyLLM2025}.

\paragraph{Extract Key Facts Using a Fixed Schema} This pattern extends the first by adding a structured interpretation. The system not only stores the raw context, but also extracts key information into a predefined format to make it easier to access and reason over. The schema can take different forms: it might be an entity map, which represents key entities (such as people, items, or places) as nodes. Each node maintains the properties of the entity (e.g., name, type), current state (e.g., location, status) and links to other entities (e.g., ``owns'', ``works with'', ``is located at''). Or, it can be event records, which works as a template that breaks an event into different aspects. In addition, it can be a task tree, where complex goals are broken down into subtasks in a hierarchical structure. A concrete example is CodeRabbit, which constructs a structured case file prior to code review, encoding cross-file dependencies, historical PR information, and team-specific rules into an explicit schema, enabling the AI to reason over the complete system context rather than isolated file changes~\cite{Sahana2025codereview}. This approach enables more effective reasoning than raw summaries by allowing the system to retrieve, analyze, and relate specific facts within an explicit schema. However, maintaining multiple layers can lead to inconsistencies, and designing good extractors to fill in these schemas reliably remains a significant challenge~\cite{du2025surveycontextawaremultiagentsystems}.

\paragraph{Progressively Compress Context into Vectors that Capture The Meaning} Apart from storing raw contexts, this approach encodes information as dense numerical vectors, known as embeddings, that reflect the meaning of the input. 
These vectors are compressible, allowing the construction of multilevel (hierarchical) memory that abstracts context at different scales.
These vectors undergo self-baking, where older embeddings are periodically summarized into compact representations (typically through pooling), or fused with existing long-term states, and re-encoded to form progressively more abstract and stable semantic memories. 
This method is compact and flexible, and is particularly useful for semantic search or matching similar queries. However, the resulting representations are not human-readable, making it challenging to edit or inspect specific parts of the memory~\cite{H_MEM2025,VectorMemory2024}.

\section{Context Usage}


\subsection{Intra-System Context Sharing}

\begin{figure}[ht]
    \centering
\includegraphics[width=1\linewidth]{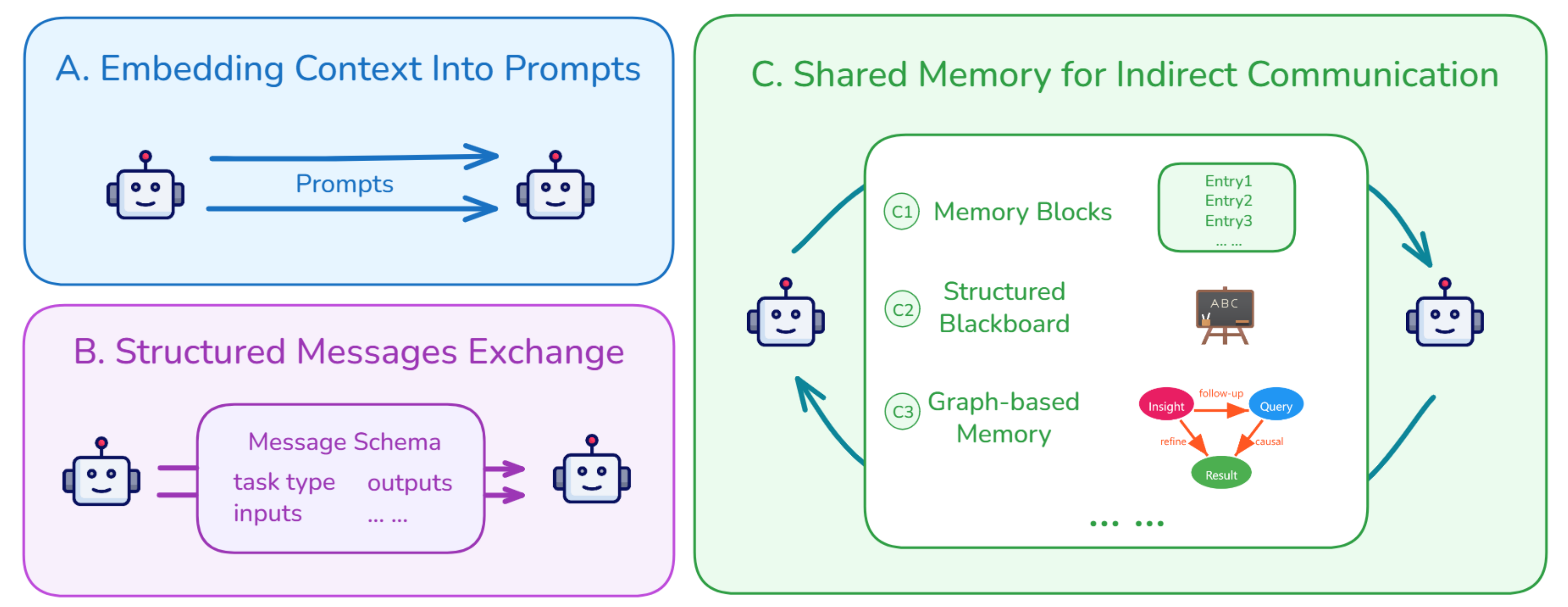}
    \caption{Common patterns of cross-agent context sharing}
    \label{fig:agent_sharev2}
\end{figure}

Modern LLM applications often comprise multiple agents, each responsible for part of a larger reasoning workflow. A practical reason why multi-agent systems work is that they allow the system to consume more tokens than a single agent could handle, effectively extending the capacity of the overall process~\cite{claude2025build}. In such systems, a key challenge arises: How can
these agents share context to achieve coherent, collaborative behavior? Unlike single-agent prompting, multi-agent systems require clear and structured ways to pass information between agents. One agent may produce a partial answer or an intermediate result, which must then be understood and used by another. To support this, the shared information must be accurate, easy to interpret, and directly usable for the next step. We identify several common patterns of cross-agent context sharing:


\paragraph{Embedding Previous Context into Prompts} This method passes information by including the previous context directly in the input prompt of the next agent. Often, the information is reformatted to be clearer. For example, one agent may summarize its thought process as plain text, which the next agent reads and continues from. This way of passing context is common in systems like AutoGPT~\cite{Team2025AutoGPT} and ChatDev~\cite{liu2025chatdevplus,qian2024chatdevcommunicativeagentssoftware}, where agents work in sequence using prompts as their communication channel.

\paragraph{Exchanging Structured Messages between Agents} Agents communicate by exchanging structured messages using a fixed format. These messages typically follow a predefined schema with fields such as task type, input data, output results, and reasoning steps. The receiving agent reads this message and continues the task. Systems like Letta~\cite{letta} and MemOS~\cite{memos} use this approach to maintain clarity and consistency when passing information between agents.

\paragraph{Using Shared Memory for Indirect Communication} Agents often communicate indirectly by reading and writing to a shared memory space, which can be implemented as a centralized external storage or as a public area within the memory of an individual agent. Instead of sending messages directly, agents leave information in this shared space, which is often organized as memory blocks, and check it later to receive updates. Each block can store a unit of context and is typically labeled with basic metadata, such as who added it, when, and what type of data it contains. Systems like MemGPT~\cite{letta} and A-MEM~\cite{amem} adopt this approach to support indirect communication and coordination between agents.
Another way is to make memory more structured: Instead of placing raw data in a general pool, information is written on a shared ``blackboard'' which is organized into segments based on topics, tasks, or goals~\cite{du2025surveycontextawaremultiagentsystems, salemi2025llmbasedmultiagentblackboardinformation}. Each agent monitors only the segments relevant to its expertise and adds or modifies the entries accordingly. This modular structure reduces confusion while allowing agents to collaborate asynchronously. 
Memory can also be organized as a graph. For example, the Task Memory Engine (TME)~\cite{ye2025taskmemory} represents an agent’s reasoning process as a task graph: Each node encodes a single step, including its input, output, and execution status, while edges capture dependencies between steps. This structure enables agents to track, reuse, and resume multi-step reasoning in a reliable and interpretable way. Similarly, G-Memory~\cite{zhang2025gmemory} models memory as a semantic graph, where nodes represent insights, user queries, and intermediate results, and edges reflect relationships such as follow-up, refinement, or causal linkage. Such graph-structured memory enables a richer representation of context and more effective reasoning across multiple steps or tasks.

\subsection{Cross-System Context Sharing}



In the context of multi-agent environments, a system can be understood as any independent platform, model, or application that maintains its own context or state for performing tasks. While different systems may vary in scale and capabilities, they each manage information within their own boundary. Sharing context across systems allows these distinct entities to access or exchange relevant information, enabling better coordination or reasoning. For example, sharing context between Cursor and ChatGPT illustrates this kind of cross-system interaction. When context is shared within a single system, it is generally easier because the components are designed to interoperate --- they mostly use compatible data formats, follow similar rules, and expect similar types of input and output. As a result, context can often be exchanged without extensive translation. In contrast, when sharing context across different systems, each system may use its own format, structure, and logic. What makes sense in one system may look completely unfamiliar to another. In this case, the challenge is to make the shared context interpretable across boundaries.






\paragraph{Use Adapters to Convert Context} Each system keeps using its own format and adds a converter to translate the context into something that the other system can read. This gives systems more freedom, but means that you have to build a separate adapter for each connection~\cite{Zarabzadeh2024Survey, VentureBeat2024MCPRelease}.

\paragraph{Using a Shared Representation Across Systems} All systems agree to use the same representation from the beginning, ensuring that each can read and write directly. This avoids the need for translation between every pair of systems and makes integration much simpler~\cite{Descope2025MCPOverview, InfoQ2025MCPConnector}. Shared representations can take different forms. Common approaches include the following:

\paragraph{Using a Shared Data Format such as JSON or an API} Systems can agree on a fixed format for the context, such as a shared JSON schema or a well-defined API. This allows each system to read and write context in a consistent way, avoiding the need for custom translation logic~\cite{Vellum2025ContextEngineering, AnalyticsVidhya2025OpenAISwarm}.

\paragraph{Sharing Context via Human-Readable Summaries} Instead of relying on formal data structures, systems can exchange short descriptions of natural language. These summaries are easy for humans to understand and can also be interpreted by language models when needed~\cite{Dobre2025NLPTasks, Verloop2025NLPChatbots}.

\paragraph{Representing Context as Semantic Vectors} Context can also be represented as numeric vectors that capture the meaning of information. This method is flexible and system independent, but often requires machine learning models to interpret correctly~\cite{Medium2025AgenticAI_MCP, DataCamp2025VectorDatabases,petrova2025semanticwebmasagentic}.

\vspace{0.5em}

Each method involves trade-offs. Standard schemas offer precision, but require strict coordination. Natural language is flexible and easy to generate, but more difficult to parse reliably. Semantic vectors are compact and generalizable, but less interpretable. The best choice depends on the systems involved and the goals of the communication.

\subsection{Context Selection for Understanding}

Even with extended context windows, LLMs remain bottlenecked by the quality of input tokens they can condition on. So an important question emerges: What subset of available context should be selected for the current step? Contexts may originate from a scratchpad, from memory, from tool definitions, or from knowledge retrieved via RAG, but not all contexts are equally useful~\cite{Lance2025Context}. Irrelevant or noisy memory fragments can distract reasoning or increase the cost of inference. Empirically, AI coding performance often decreases when context windows exceed roughly 50\% fullness, suggesting that both excessive and insufficient context can degrade effectiveness~\cite{Addy2025tips}. Without robust filtering, agents risk being overwhelmed by their own memory, a modern form of ``context overload''. Thus, effective context selection becomes a form of \textit{attention before attention} --- choosing what deserves to be paid attention.




With the rise of LLM-powered agents, the context engineering 2.0 era shifted toward dynamic, goal-driven interaction. Context filtering became an adaptive process, dynamically selecting the information that is most relevant to the current intent of the user~\cite{DailyLLM2025, LayeredMemory2025}. We identify several factors that should be considered during memory selection:

\paragraph{Semantic Relevance} It refers to selecting memory entries that are most similar in meaning to the current query or objective. This is typically implemented via vector-based retrieval: both query and candidate entries are encoded as dense embeddings, and a nearest-neighbor search (e.g., FAISS or other approximate nearest-neighbor methods) retrieves items by similarity, allowing semantically related items to be found even without exact keyword matches~\cite{johnson2019billion}. This approach is commonly used in retrieval-augmented generation (RAG) pipelines and systems such as Letta~\cite{letta, VectorMemory2024, H_MEM2025}.

\paragraph{Logical Dependency} Logical dependency refers to cases where the current task directly relies on information produced by previous steps, such as prior planning decisions, outputs from tools, or reasoning chains. Systems such as MEM1 address this by explicitly recording reasoning traces during execution: whenever a new step is generated, MEM1 stores it in a memory slot and links it to the earlier steps it depends on (e.g., prior analyses or tool outputs). Over time, this forms a structured dependency graph across memory entries. When a new query arises, MEM1 can traverse this graph to retrieve only the context that lies in the relevant dependency chain, ensuring that memory reflects not just surface relevance but also the logical flow of the task~\cite{mem1, HiMemFormer2024, HierarchicalMemoryTransformer2025}.

\paragraph{Recency and Frequency} Items that were used recently or accessed often are more likely to be retrieved again. This is based on simple heuristics: if something was useful before, it might be useful again. To implement this, systems typically assign higher priority to newer memory entries and gradually lower the priority of older ones, so that outdated information naturally becomes less influential. At the same time, entries that are repeatedly accessed accumulate a higher importance, ensuring that commonly referenced information remains readily available. By balancing these factors, the memory pool evolves to favor fresh and important contexts~\cite{letta, LayeredMemory2025}.

\paragraph{Overlapping information} If multiple pieces of information convey the same meaning, the older or less detailed ones can be filtered out~\cite{du2025surveycontextawaremultiagentsystems, jiao-etal-2024-text2db}. In modern systems, this process is increasingly handled through active memory management, where the system does not merely detect similarity passively but actively decides when to merge, update, or remove redundant entries to maintain a concise and relevant memory pool~\cite{MemoryR1}.

\paragraph{User Preference and Feedback} Over time, AI agents can adapt to a user’s habits: learning what types of information the user tends to value. Some systems, such as self-evolution memory~\cite{selfevolution, HierarchicalMemoryTransformer2025}, track how users interact with information and use it to adjust the importance or weight of memory entries.

\vspace{0.5em}

A primary consideration in memory selection is \textbf{relevance}. This can be assessed by several factors, including \textbf{semantic similarity}, \textbf{logical dependency}, \textbf{recency} (the timeliness of the information), and \textbf{frequency of mention}. For example, many systems employ similarity scores to compare current input with stored content, assigning higher relevance to information with greater similarity. Alternatively, explicit tags or metadata can be incorporated to indicate the importance or function of specific data, such as marking certain events as milestones or highlighting key facts. In addition to relevance, it is important to \textbf{minimize redundant information} and \textbf{adapt to user habits}. By integrating these criteria, systems are able to retain the most pertinent data, reduce unnecessary noise, and enhance the efficiency of context selection.


\paragraph{Common Filtering Strategies}

Taking these factors into account, the systems adopt different filtering strategies. In RAG pipelines, the first step is to segment the source into manageable chunks, which may be done with simple strategies such as fixed line or token windows, or with more structured approaches like AST-based segmentation that respects function, class, or module boundaries and preserves semantic coherence. Retrieval then proceeds in several distinct ways. Semantic retrieval often relies on an embedding-based search, which selects fragments according to the vector similarity to the query~\cite{VectorMemory2024}. Non-semantic retrieval can be as direct as Grep, which uses string or regex matching without interpreting meaning. Structured retrieval uses knowledge graphs, drawing on entities and dependency relations (e.g., function-call graphs) to connect information across files and modules~\cite{wen2025effectiveefficientschemaawareinformation}. Because these methods frequently return overlapping or noisy candidates, many systems introduce a reranking stage (sometimes powered by LLMs) to refine relevance, trading off accuracy improvements against efficiency, with implementations (e.g., Windsurf) varying in their design choices~\cite{Varun2025Windsurf}.

\subsection{Proactive User Need Inference}

Most current uses of context are reactive, yet users often find it difficult (or unreasonable) to fully articulate their needs and preferences. To bridge this gap, context engineering should enable agents to act \textit{proactively}: to infer latent user needs, preferences, and goals that are not explicitly stated, and to initiate helpful interactions accordingly. This is analogous to human assistants who learn over time that ``user prefers visual summaries'', or that ``evening hours are best for brainstorming''. Such insights are not predefined; they must be mined, abstracted, and validated through the accumulated interaction context. In era 2.0, we identify several common forms of proactive preference mining:




\paragraph{Learning and Adapting to User Preferences}
Modern AI agents learn and adapt to user preferences primarily by analyzing conversation history and stored personal data such as user documents, notes, and past interactions, in order to identify patterns in communication style, interests, and decision-making approaches~\cite{pan2025memoryconstructionretrievalpersonalized, sun2024largelanguagemodelsempower}. In many cases, these insights are consolidated into evolving user profiles that guide more personalized interactions.
They also learn from indirect signals by observing how users respond to suggestions, whether they continue conversations, and their apparent satisfaction with completed tasks, incorporating these observations into future interactions~\cite{openai2024chatgpt}.
In addition, sometimes agents actively explore preferences by directly asking users about their likes and dislikes or presenting multiple options during conversations to gather explicit feedback~\cite{ryu2025syntheticdialoguegenerationinteractive}.

\paragraph{Inferring Hidden Goals from Related Questions}
The system can infer hidden goals by analyzing the sequence of user queries. For instance, if a user asks about Python decorators and then about performance tuning, this may reflect a broader goal of improving software design efficiency. To support such inference, systems can encode the current context and put it into an LLM to predict users' potential needs~\cite{anthropic2025claude}. Moreover, chain-of-thought reasoning techniques enable systems to perform multi-step logical deduction, inferring deep intentions from surface-level user inputs~\cite{yao2025tree}. Recognizing these patterns, the system can offer more goal-aligned assistance.

\paragraph{Proactively Offering Help Based on User Struggles}
The system detects when the user may be stuck, such as showing hesitation or trying many alternatives, and proactively offers useful tools like visualizations or checklists~\cite{chen2025needhelpdesigningproactive}. These interventions aim to improve the quality of decision without asking the user.

\subsection{Lifelong Context Preservation and Update}



As noted in the abstract, individuals are fundamentally shaped by their interactions with other entities --- in other words, by their contexts. Given this shift, the central challenge naturally arises: Once context takes on a lifelong form, how should it be preserved and updated in a way that remains coherent, adaptive, and usable for both humans and machines? The continuous preservation and updating of lifelong context demands more than scalable storage: It requires the design of a memory system that is dynamic, semantically robust, and temporally aware. This raises a multifaceted technical challenge, where naive extensions of current memory architectures collapse under the scale, fluidity, and complexity of real-world context.

Here is the storyline: When ``normal'' context engineering becomes lifelong context engineering, a problem arises: How to implement \textbf{reliable storage mechanisms} that maintain semantic consistency? Moreover, mere storage is insufficient; as data scales to massive volumes, the system must be capable of accurately \textbf{processing and managing} this information. Once a new design is proposed, an end-to-end \textbf{evaluation} framework becomes essential to validate both its correctness and performance. However, such mechanisms are still largely underdeveloped. Once the complete pipeline is established, maintaining its \textbf{stability} becomes paramount. Each stage within this workflow presents distinct challenges, particularly in the domain of lifelong context systems.

\paragraph{Challenge I: Storage Bottlenecks}

The first challenge is how to retain as much relevant context as possible under strict resource constraints. We currently lack a unified solution: How can we preserve as much context as possible, ensuring that all of my contexts can be effectively retained without loss? What kind of infrastructure or interface would facilitate recording our context to the maximum extent? And how can storage systems simultaneously support high-compression, high-precision retrieval, and low-latency access at scale~\cite{xing2025structuredmemorymechanismsstable, ahn2025hemahippocampusinspiredextended}?

\paragraph{Challenge II: Processing Degradation}

Another challenge arises from the collapse of attention mechanisms at scale. Most transformer-based models rely on global attention, whose \(O(n^2)\) complexity leads to rapidly increasing inference latency, high GPU memory usage, and slower I/O throughput. These resource bottlenecks make it impractical to handle large contexts in real time.
Meanwhile, the quality of reasoning deteriorates. As attention becomes thinner across a longer input, the model struggles to maintain focus on relevant information, and it struggles to capture long-distance dependencies~\cite{THAI2023107518, rabe2022selfattentiondoesneedon2}. In addition, retrieval systems become overwhelmed by many semantically similar but irrelevant pieces of information, often returning distractions instead of useful evidence~\cite{jeong2024adaptive}. Furthermore, as the volume of context grows, inconsistencies and conflicts between retrieved segments become harder to detect and reconcile.
All of these issues lead to fragile reasoning when dealing with very large contexts.

\paragraph{Challenge III: System Instability}

As memory accumulates over time, even small mistakes can affect more parts of the system. Errors that once had limited impact may now spread widely, leading to unexpected or unstable behavior. Without clear boundaries or validation mechanisms, the system becomes more difficult to manage, especially in tasks that run for a long time or require high safety. In such cases, it might make the system more fragile instead of increasing reliability~\cite{meng2025preserving, jacobs2025semantic}.

\paragraph{Challenge IV: Difficulty of Evaluation}

As memory accumulates, it becomes harder to tell whether the system is reasoning correctly. Most benchmarks today only test whether the system can retrieve information, but do not check whether the information is still relevant, accurate, or helpful. Systems rarely include features to check for contradictions, undo wrong updates, or trace the reasoning steps that led to a conclusion. As more decisions depend on longer memory chains, it becomes increasingly difficult to inspect how the system reached a specific answer. This lack of visibility makes it harder to trust or improve the system, especially for lifelong context engineering~\cite{zheng2025lifelongagentbenchevaluatingllmagents}.

\paragraph{Toward a Semantic Operating System for Context}

The challenge of lifelong context engineering can no longer be addressed by simply ``expanding the context window'' or ``improving retrieval accuracy''. It demands the construction of a \textbf{semantic operating system} capable of growing over time, much like a human mind. On the one hand, such a system must support large-scale, efficient semantic storage as its own memory bank, and exhibit truly human-like memory management abilities: actively adding, modifying, and forgetting knowledge. On the other hand, it calls for \textbf{novel architectures} to replace Transformers' flat temporal modeling, thereby enabling more powerful long-range contextual reasoning and dynamic adaptation. Crucially, the system should be able to explain itself by tracing, correcting, and interpreting each step in its reasoning chain, thus improving trust and reliability in practical and safety-critical scenarios. This approach highlights a fundamental principle of context engineering and reflects a paradigm shift: context is no longer passively accumulated, but is actively managed and evolved as a core element for cognition.

\subsection{Emerging Engineering Practices}

\paragraph{KV Caching} 
Under emerging engineering practices, the use of key-value (KV) caching has become central to the efficient deployment of agents. KV caching works by storing the attention states (keys and values) of past tokens so they do not need to be recomputed when new tokens are generated. As a result, the cache hit rate strongly affects both latency and cost. To improve hit rate, several practices are proving essential: first, keeping prefix prompts stable, since even minor variations such as timestamps at the beginning of the system prompt can invalidate the entire cache; second, enforcing append-only and deterministic updates, because altering or inconsistently serializing past content breaks reuse; and third, in cases where serving frameworks do not support automatic incremental prefix caching, it is necessary to insert cache checkpoints manually and place them carefully~\cite{manus2025context}. In addition, cache warm-up is often employed to further enhance efficiency. A common approach is predictive loading (prefetch or speculative loading), in which the system anticipates which contexts are likely to be needed next and loads them into the cache in advance~\cite{Lex2025Transcript}. These techniques show that efficiency is increasingly determined by the way context is managed.

\paragraph{Tool Designing} 
The important factors in tool design are description and scale. For description, tools need precise purposes and clear definitions. Vague or overlapping descriptions often cause failures, while well-structured ones reduce ambiguity and improve reliability. Models can also refine these descriptions and act as prompt engineers, enabling self-improvement~\cite{claude2025build}.
For scale, large tool sets make agents less reliable, and dynamically loading tools during interaction often breaks KV-cache consistency and confuses references to earlier actions. Empirical observations suggest that excessively large tool sets can degrade performance, as overlapping tool descriptions and increased choice complexity make selecting the correct tool more difficult. For DeepSeek-v3, performance declined beyond 30 tools and was nearly guaranteed to fail beyond 100~\cite{dbreunig2025fix}. A more robust approach is to keep the tool list stable and enforce constraints at the decoding level, for example, by masking token logits to block invalid choices. This practice preserves efficiency while reducing errors caused by changing the action space~\cite{manus2025context}.

\paragraph{Context Contents}

An agent should not hide an agent’s mistakes; retaining errors in the context allows the model to observe its failures, which is crucial for learning corrective behavior and improving overall performance~\cite{manus2025context}.
In agent settings, traditional few-shot prompting can be counterproductive. When the context contains largely similar past action-observation pairs, the model tends to repeat prior actions, simply following the patterns it sees. To mitigate this, Manus introduces small, structured variations in actions and observations, such as alternative serialization templates, varied phrasing, or subtle changes in order and formatting. These controlled perturbations break repetitive patterns and help refocus the model’s attention, improving robustness and reducing the risk of overgeneralization~\cite{manus2025context}.

\paragraph{Multi-agent Systems} 
Claude’s experience suggests that effective multi-agent work depends on a few recurring practices. The lead agent, or lead researcher, breaks queries into subtasks and assigns them with clear goals, outputs, tool guidance, and boundaries, as vague instructions lead to confusion or gaps. Because agents struggle to judge workload, prompts can include simple adjustment rules relevant to query complexity, for example, simple tasks may need 1 agent to complete, while harder ones may need more. Search strategies work best when they move from broad exploration to focused analysis, and extended thinking mode, where agents explicitly write down their reasoning process, enhances overall accuracy and efficiency~\cite{claude2025build}.

\paragraph{Tricks} 

In executing complex tasks, many systems maintain a todo.md file listing subgoals, updating it and marking items as completed as the task progresses. However, models can lose track of earlier objectives in long tasks. A practical solution is to recite these goals in natural language when updating the todo list, integrating them in the recent context of the model to keep the key objectives within its immediate attention~\cite{manus2025context}.

\section{Applications}

\subsection{CLI}

When using an AI agent, developers often require sustained, project-oriented context support. Google’s \emph{Gemini CLI} offers a representative case of how context can be engineered. A central mechanism is the \emph{GEMINI.md} file, a Markdown specification that records the project background, role definitions, required tools and dependencies, coding conventions, etc. The contexts are organized through the file system hierarchy: \emph{GEMINI.md} files can exist in the user's home directory, in the project root, or within subdirectories, enabling both inheritance and isolation of information~\cite{schmid2025geminicheatsheet}. 
For collection, the CLI gathers two types of context: at startup, it automatically loads static information such as system prompts, the current project environment, and the ancestor or descendant \emph{GEMINI.md} files; during interaction, it incrementally accumulates dynamic context from the ongoing dialogue history~\cite{irani2025gemini, sii-cli-2025}.
For management, the file system itself functions as a lightweight database, with mechanisms to compress context by replacing long interaction histories with AI-generated summaries. These summaries follow predefined formats that preserve key aspects of the dialogue (e.g., overall goal, key knowledge, file system state, recent actions, and current plan), ensuring consistency. Community discussions have also suggested extending this process with human refinement to support collaborative management of context~\cite{irani2025geminiclilab, gemini_cli, xiao2025limiagency}.

\subsection{Deep Research}

Deep research agents aim to assist users in addressing open-ended, knowledge-intensive queries, such as reasoning tasks involving multiple events and intertwined relationships. A representative example is Tongyi DeepResearch, which operates in four main steps: it searches the web based on the user’s query, extracts key information from relevant pages, generates new sub-questions to guide further search, and finally integrates evidence from multiple sources into coherent answers~\cite{tongyidr}. This cycle often continues for many rounds until uncertainty is reduced and a complete evidence chain is formed. Unlike short-turn conversational agents, deep research faces the challenge of extremely long interaction histories: directly appending all observations, thoughts, and actions quickly exceeds the context window. To overcome this limitation, Tongyi DeepResearch adopts systematic context engineering. During exploration, the agent periodically invokes a specialized summarization model to compress accumulated history into a compact reasoning state, which not only preserves critical evidence but also highlights missing information and next-step directions. Subsequent searches and reasoning are then grounded in these compressed ``context snapshots'' rather than the complete raw history. In this way, the system establishes a clear context lifecycle: from collecting and accumulating information, to periodic compression and abstraction, and then to reasoning and reuse based on summaries, allowing it to break through context constraints and achieve scalable, long-horizon research capabilities~\cite{wu2025resumunlockinglonghorizonsearch}.

\subsection{Brain-Computer Interfaces}

Brain-Computer Interfaces (BCIs) offer a novel pathway for context engineering by enabling more advanced context collection~\cite{tang2023flexible}. Unlike traditional methods that rely on language input, BCIs can directly capture neural signals, which introduces two distinct advantages. First, they allow the collection of richer contextual dimensions, such as attention levels, emotional states, or cognitive load --- factors that are often difficult to observe through external behavior alone. Second, they provide a more convenient way of collecting context, reducing the need for explicit user actions, and enabling more immediate input through neural activity~\cite{liu2025memristor, edelman2024non}. Although current techniques offer only a coarse understanding of brain signals and issues such as noise and instability remain significant challenges, BCIs highlight a direction in which context engineering may evolve: extending context collection beyond external environments to the user’s internal cognitive state~\cite{wu2023affective}.

\section{Challenges and Future Directions}

Although we have outlined a historical and conceptual framing of context engineering, a number of open challenges remain. Below we highlight several key issues and sketch possible directions for future exploration.

\paragraph{Context collection remains limited and inefficient}
Most current agent systems still rely on explicit user input to obtain context, which is both cumbersome and inefficient~\cite{zhao2024retrieval}. In
addition, users are sometimes unable to articulate their intentions clearly, leaving important contextual information underspecified. Progress requires more natural and multimodal methods of context collection, together with models that can better infer user needs and fill in missing contexts~\cite{chen2024evolution}. One promising line of work is brain–computer interfaces, which aim to collect user state and intent more efficiently than explicit articulation. Such advances suggest that richer forms of context collection may eventually overcome the inherent limitations of text-based input~\cite{liu2025memristor, edelman2024non}.

\paragraph{Storage and management of large-scale context}
As interactions accumulate, the size and complexity of the context grow rapidly. A key challenge lies in deciding how context should be stored and organized and how to organize it in a way that remains scalable while supporting effective selection and retrieval. Without careful design, large-scale context can become cumbersome and difficult to use for downstream tasks~\cite{adnan2025longcontext}.

\paragraph{Limited model understanding of context}
Current systems do not possess the same level of contextual understanding as humans. For example, they struggle with complex logic and with relational information in images~\cite{yang2023survey}. As mentioned in this paper, lower machine intelligence leads to greater ``effort'' in context engineering. As a result, much of the available context is not fully understood or utilized. Future work should focus on strengthening models’ abilities in semantic reasoning, logical interpretation, and multimodal alignment, enabling systems to better understand context and reduce reliance on human-driven context engineering~\cite{pan2023large}.

\paragraph{Performance bottlenecks with long context}
The processing of long contexts remains a central challenge. Transformer-based architectures suffer from quadratic complexity, making them inefficient as the context grows. Recent alternatives like \textit{Mamba}~\cite{gu2023mamba} and its variants such as \textit{LongMamba}~\cite{ye2025longmamba} have improved efficiency and scalability, but still show weaknesses in long-context understanding. For example, when input length far exceeds training length, or when relational and logical dependencies span the entire context. \textsc{LOCOST}~\cite{bronnec2024locoststatespacemodelslong} can handle documents of hundreds of thousands of tokens, but still lags behind transformers in certain tasks that require fine-grained reasoning over very long spans. Moving forward, there is a clear need for new architectures that can handle much longer contexts efficiently while also providing stronger and more reliable understanding, rather than treating long contexts merely as extended input.

\paragraph{Selecting relevant and useful context}
Not all available context contributes to the task at hand. Although current systems already employ relevance estimation and filtering mechanisms, their performance remains limited: useful signals may be missed, and noisy or redundant information often persists. A key research direction is to develop more precise and adaptive context selection strategies that continuously refine what to keep, discard, or emphasize, ensuring that the retained context remains closely aligned with the objectives of the task~\cite{VectorMemory2024, Varun2025Windsurf}.

\paragraph{Digital Presence} 
Karl Marx once wrote that ``the human essence is the ensemble of social relations''~\cite{marx1845theses}. In the era of context-centric AI, this idea takes on a new computational meaning: individuals are increasingly defined not by their physical presence or conscious activity, but by the digital contexts they generate: their conversations, decisions, and traces of interaction. These contexts can persist, evolve, and even continue to interact with the world through AI systems long after the departure of a person. The human mind may not be uploaded, but the human context can --- turning context itself into a lasting form of knowledge, memory, and identity.

\section{Conclusion}

In this paper, we explore the context of context engineering, arguing that it is not a sudden invention of the LLM era but a long-evolving discipline shaped by the progressive intelligence of machines. By tracing its historical phases and by outlining design considerations that govern its practice, we highlight how the core challenge lies in bridging human intent and machine understanding under varying levels of entropy. Our proposed trajectory suggests a gradual human disengagement from explicit context management, as increasingly intelligent machines take on greater responsibility for interpreting, reasoning, and even constructing context. Looking ahead, as machine understanding approaches and potentially surpasses human cognition, culminating in a possible ``god’s eye view'' of our intentions, AI systems may not only comprehend us, but also illuminate and expand our understanding of ourselves.

\section*{Acknowledgement}
Our sincere thanks go to Xiangkun Hu from Amazon for his early discussions. We are also deeply thankful to Shijie Xia from Shanghai Jiao Tong University for his valuable feedback.


\end{document}